%% file: main.tex
\documentclass[sigconf,screen]{acmart}
\acmConference[WWW 2023]{The Web Conference}{30 April - 4 May 2023}{United States}

\AtBeginDocument{%
  \providecommand\BibTeX{{%
    \normalfont B\kern-0.5em{\scshape i\kern-0.25em b}\kern-0.8em\TeX}}}

\usepackage{calligra}
\usepackage{algorithm}
\usepackage{algorithmicx}
\usepackage{algpseudocode}

\usepackage{xcolor,colortbl}
\usepackage{xspace}
\usepackage{booktabs}
\usepackage{moreverb}
\usepackage{fontenc}
\usepackage{amsmath}
\usepackage{fancybox}
\usepackage{color}
\usepackage{colortbl}
\usepackage{array}
\usepackage{multirow}
\usepackage{multicol}
\usepackage{listings}
\usepackage{makecell}
\usepackage{graphicx}
\usepackage{setspace}
\usepackage[multiple]{footmisc}
\usepackage{wasysym}
\usepackage{bigstrut}

\usepackage{balance}
\usepackage{csvsimple}
\usepackage{longtable}
\usepackage{caption}
\usepackage{url}
\usepackage{lscape}
\usepackage{rotating}
\usepackage{tikz}
\usepackage{enumitem}
\usepackage{subfigure}

\usepackage{marvosym}
\usepackage{pifont}
\usepackage{courier}
\usepackage{afterpage,natbib,lipsum}

\algnewcommand\algorithmicforeach{\textbf{for each}}
\algdef{S}[FOR]{ForEach}[1]{\algorithmicforeach\ #1\ \algorithmicdo}

\newcolumntype{L}[1]{>{\raggedright\let\newline\\\arraybackslash\hspace{0pt}}m{#1}}
\newcolumntype{C}[1]{>{\centering\let\newline\\\arraybackslash\hspace{0pt}}m{#1}}
\newcolumntype{R}[1]{>{\raggedleft\let\newline\\\arraybackslash\hspace{0pt}}m{#1}}

\definecolor{codegreen}{rgb}{0,0.6,0}
\definecolor{codered}{rgb}{1,0,0}
\definecolor{codegray}{rgb}{0.5,0.5,0.5}
\definecolor{codepurple}{rgb}{0.58,0,0.82}
\definecolor{backcolour}{rgb}{0.95,0.95,0.92}
\definecolor{lightgray}{gray}{0.9}

\lstdefinestyle{mystyle}{
    commentstyle=\color{codegreen},
    keywordstyle=\color{magenta},
    numberstyle=\small\color{black},
    stringstyle=\color{codepurple},
    basicstyle=\scriptsize\ttfamily,
    breakatwhitespace=false,
    breaklines=true,
    captionpos=b,
    keepspaces=true,
    showspaces=false,
    showstringspaces=false,
    showtabs=false,
    tabsize=2
}

\lstdefinelanguage{diff}{
  morecomment=[f][\color{blue}]{@@},     
  morecomment=[f][\color{red}]-,         
  morecomment=[f][\color{codegreen}]+,       
  morecomment=[f][\color{red}]{---}, 
  morecomment=[f][\color{codegreen}]{+++}
}

\lstdefinelanguage{text}{
  breaklines=false
}

\setlist{noitemsep} 

\definecolor{darkpastelred}{rgb}{0.76, 0.23, 0.13}
\definecolor{ao(english)}{rgb}{0.0, 0.5, 0.0}

\definecolor{darkpastelred}{rgb}{0.76, 0.23, 0.13}
\definecolor{ao(english)}{rgb}{0.0, 0.5, 0.0}

\definecolor{yellow}{RGB}{255,255,153}
\definecolor{grey}{RGB}{224,224,224}

\newboolean{showcomments}
\setboolean{showcomments}{true}
\ifthenelse{\boolean{showcomments}}
 { \newcommand{\mynote}[2]{
      \fbox{\bfseries\sffamily\scriptsize#1}
        {\small$\blacktriangleright$\textsf{\emph{#2}}$\blacktriangleleft$}}}
        { \newcommand{\mynote}[2]{}}

\definecolor{DarkOrange}{rgb}{0.8,0.3,0.0}
\definecolor{DarkCyan}{rgb}{0.0, 0.55, 0.55}
\definecolor{DarkCyel}{rgb}{1.0, 0.49, 0.0}
\definecolor{yellow-green}{rgb}{0.6, 0.8, 0.2}

\newcolumntype{?}{!{\vrule width 1pt}}

\newcommand{\etal}{\emph{et~al.}\xspace}
\newcommand*{\ie}{i.e., }
\newcommand*{\eg}{e.g., }

\definecolor{turquoise}{rgb}{0.1, 0.75, 0.6}

\setcopyright{none}

\renewcommand\footnotetextcopyrightpermission[1]{} 

\copyrightyear{2023}
\acmYear{2023}
\setcopyright{rightsretained}
\acmConference[ESEC/FSE '23]{The ACM Joint European Software Engineering Conference and Symposium on the Foundations of Software Engineering 2023}{November 11-17, 2023}{San Francisco, California, United States}

\begin{document}
\title[FITNESS: A Causal De-correlation Approach for Mitigating Bias in Machine Learning Software]{
FITNESS: A Causal De-correlation Approach for Mitigating Bias in Machine Learning Software
}

\author{Ying Xiao}
\email{12150075@mail.sustech.edu.cn}
\affiliation{%
   \institution{Southern University of Science and Technology}
 	\country{China}
}

\author{Shangwen Wang}
\email{wangshangwen13@nudt.edu.cn}
\affiliation{%
  \institution{National University of Defense Technology}
  \country{China}
}

\author{Sicen Liu}
\email{11910338@mail.sustech.edu.cn}
\affiliation{%
   \institution{Southern University of Science and Technology}
 	\country{China}
}

\author{Dingyuan Xue}
\email{11910213@mail.sustech.edu.cn}
\affiliation{%
   \institution{Southern University of Science and Technology}
 	\country{China}
}

\author{Xian Zhan}
\email{chichoxian@gmail.com}
\affiliation{%
  \institution{Huawei}
 	\country{China}
}

\author{Yepang Liu}\authornote{Corresponding author.}
\email{liuyp1@sustech.edu.cn}
\affiliation{%
   \institution{Southern University of Science and Technology}
 	\country{China}
}

\input{0.abstract}


%


 
\maketitle


\input{1.intro}
\input{2.background}

\input{3.approach}
\input{4.evaluation}
\input{5.results}
\input{6.discussion}
\input{7.conclusion}
\input{8.availability.tex}

\begin{acks}
This work was supported by the NATURAL project, which has received funding from 
the European Research Council under the European Union’s Horizon 2020 research 
and innovation program (grant No. 949014).
Kui Liu was also supported by the National Natural Science Foundation of China (Grant No. 62172214), the Natural Science Foundation of Jiangsu Province, China (Grant No. BK20210279), and the Open Project Program of the State Key Laboratory of Mathematical Engineering and Advanced Computing (No. 2020A06).
%
\end{acks}

\balance
\bibliographystyle{ACM-Reference-Format}
\bibliography{references}

\end{document}

%% file: 0.abstract.tex
\begin{abstract}
Software built on top of machine learning algorithms is becoming increasingly prevalent in a variety of fields, including college admissions, healthcare, insurance, and justice. The effectiveness and efficiency of these systems heavily depend on the quality of the training datasets. Biased datasets can lead to unfair and potentially harmful outcomes, particularly in such critical decision-making systems where the allocation of resources may be affected. This can exacerbate discrimination against certain groups and cause significant social disruption.
To mitigate such unfairness, a series of bias-mitigating methods are proposed.
Generally, these studies improve the fairness of the trained models to a certain degree but with the expense of sacrificing the model performance.
In this paper, we propose FITNESS, a bias mitigation approach via de-correlating the causal effects between sensitive features (\eg the sex) and the label.
Our key idea is that by de-correlating such effects from a causality perspective, the model would avoid making predictions based on sensitive features and thus fairness could be improved. 
Furthermore, FITNESS leverages multi-objective optimization to achieve a better performance-fairness trade-off.
To evaluate the effectiveness, we compare FITNESS with 7 state-of-the-art methods in 8 benchmark tasks by multiple metrics.
Results show that FITNESS can outperform the state-of-the-art methods on bias mitigation while preserve the model's performance:
it improved the model's fairness under all the scenarios while decreased the model's performance under only 26.67\% of the scenarios. 
Additionally, FITNESS surpasses the Fairea Baseline in 96.72\% cases, outperforming all methods we compared.




\end{abstract}

%% file: 1.intro.tex
\section{INTRODUCTION}
\label{sec:intro}


Machine learning (ML)-based autonomous decision-making systems are critical components of modern software service ecosystems. They are widely adopted in various application scenarios such as junk mail sorting, personalizing advertisements recommendations, autonomous vehicles, and so on \cite{kotary2022end, tsioutsiouliklis2022link, li2021user, lecun2015deep, janai2020computer, schwarting2018planning}. The application for such systems, however, requires a large-scale dataset for the adequate training of the model, and it is challenging to collect enough high-quality and unbiased data. Furthermore, the bias relating to natural human properties (\eg race, gender, age) can lead to unwanted discrimination in human-related tasks, including college admission, health care, insurance and justice, which can do harm to people's critical benefit \cite{zhang2020machine, chen2022maat, mehrabi2021survey}. As a result, how to mitigate data bias or train a discrimination-free model from a biased dataset is becoming a crucial problem in both AI and Software Engineering domains.

Fairness is such a property that could be threatened by the bias in the datasets.\footnote{We use ``bias'' and ``unfairness'' interchangeably, as the opposite of ``fairness''.} 
From the machine learning perspective, fairness is a non-functional property, but it is a fundamental right in the human society \cite{kuipers1984causal, zhang2021ignorance, chen2022maat}.
Obviously, people are not willing to be treated differently for different skin colors, sexes, and nationalities by others or the ML-enabled systems \cite{zhang2020machine, mehrabi2021survey, madaio2020co}.
Despite that, if the collected dataset is imbalanced, systems trained on it would give privilege to a certain type of people (\eg men against women), leading to the unfairness in the prediction results of the systems.
A typical example is that Amazon scraped an automated recruiting tool in 2018 as it was found to discriminate against women~\cite{reutersAmazonScraps, chakraborty2021bias}.


Recent years have witnessed a growing number of approaches being proposed to address the unfairness of ML-based models, from the Artificial Intelligence, Human Computation and Software Engineering communities \cite{hube2019understanding, sheng2019machine, hettiachchi2021investigating, mehrabi2021survey, zhang2020machine,  madaio2020co, li2021user,chen2022maat}.
Generally, existing approaches use either over-sampling or under-sampling techniques on the dataset, with the aim of balancing the data used for training \cite{chakraborty2021bias, chakraborty2020fairway, zemel2013learning, kamiran2012data}. 
However, such approaches may have specific weaknesses. For example, over-sampling could lead to the overfitting of the model to the over-sampled data \cite{mohammed2020machine} while under-sampling may provide inadequate data for model learning.
Consequently, state-of-the-art approaches could alleviate the model bias to a certain degree, but usually with the expense of a sharp drop in model performance, which is an important functional property reflecting the accuracy of the prediction from the model \cite{zhang2018mitigating, mehrabi2021survey, zhang2020machine, chakraborty2021bias, chakraborty2020fairway, chen2022maat}.
For instance, after applying Fair-SMOTE to protect the sex feature, the accuracy of a logistic regression classifier decreases from 0.83 to 0.73 in predicting whether someone is high-income in the Adult Census Income dataset \cite{misc_adult_2, chakraborty2021bias}. 

To tackle the unfairness problem, we propose a new causal de-correlation based approach named FITNESS (\textbf{F}or m\textbf{I}tigating da\textbf{T}a a\textbf{N}d mod\textbf{E}l bia\textbf{S}e\textbf{S}).
Our basic insight is that the unfair predictions from a model could be largely related to the sensitive features (\eg the sex and race) whose distributions are imbalanced in a dataset.
For instance, in a dataset to predict the income \cite{misc_adult_2}, around 80\% of the high-income people are actually male, and thus the difference between the contribution of males and females to a high-income prediction 
could be very significant \cite{chakraborty2021bias, peng2021xfair}. 
That is, models trained on such a dataset could prefer to predict a male as high-income instead of a female.
Therefore, the core idea of our approach is to eliminate the causal difference among various sensitive feature values for specific data labels to alleviate the model bias better. Specifically, we model the decision-making process as a structural causal model \cite{pearl2009causality} and take the causal intervention to quantitatively calculate the correlation between the sensitive feature values and the data labels. And then, we eliminate the causal difference by de-correlation, which involves mutating biased data instances. Consequently, we formulate the fairness problem as a search process for which data instances should be mutated to obtain an unbiased training dataset. To balance model performance and fairness better, we apply the well-known Multi-Objective Optimization (MOO) \cite{censor1977pareto} that takes into consideration of both performance and fairness metrics in the search process.

To evaluate FITNESS, we conduct experiments with 8 decision-making tasks using 3 different machine learning algorithms (\ie LR, RF, SVM) and four benchmark datasets from different domains (\eg finance, justice, and economy). 
Our experiments show that among all scenarios, FITNESS improves 100\% models' fairness and only decreases 26.67\% models' performance, which outperforms the state-of-the-art method MAAT \cite{chen2022maat} that improves 94.44\% models' fairness and decreases 38.89\% models' performance.
Regarding the state-of-the-art benchmarking tool (Fairea Baseline) \cite{hort2021fairea}, FITNESS improved the fairness and performance of 34.90\% models simultaneously and outperforms 96.72\% Fairea Baseline, surpassing all the existing methods we compared.

In summary, this paper makes the following contributions: 
\begin{itemize}[leftmargin=*]
    \item We first tackle the fairness problem from a causality perspective by de-correlating the sensitive feature values and labels.
    \item We propose a pre-processing approach named FITNESS that incorporates causal intervention and multi-objective optimization to achieve the performance-fairness trade-off.
    \item We conducted large-scale experiments and the results demonstrate FITNESS significantly surpasses the state-of-the-art.
    
    \item We also release all data and code \footnote{\label{source}\url{https://doi.org/10.5281/zenodo.7608487}} to the research community to facilitate follow-up studies or to replicate and extend FITNESS. 
\end{itemize}




%% file: 2.background.tex
\section{PRELIMINARIES}
\label{sec:bg}
In this section, we introduce some background knowledge and related works to this study.

\subsection{Background}
Integrating machine learning models into software is becoming more prevalent as a way to improve effectiveness and efficiency by leveraging historical data \cite{zhang2020machine, zhang2021ignorance, chen2022maat}. Examples include using machine learning software by employers to automatically select the most suitable resumes, by doctors to aid in diagnoses, and by banks to evaluate loan applications \cite{chakraborty2021bias, chakraborty2020fairway, chen2022maat}. Although machine learning software improves social efficiency, it faces a significant risk of fairness bugs which can exacerbate discrimination and inequality \cite{mehrabi2021survey, assembly1948universal}. This is because the training dataset substantially influences the performance of the machine learning model while the dataset from the real world is usually imbalanced in the features class and labels class \cite{lecun2015deep}. Thus, model discrimination and bias are prevalent in actual application. Because of the ethical duty of software engineers to develop fair and discrimination-free software, research on fairness problems in machine learning software is needed and meaningful. 

As the problem of bias in machine learning algorithms is not only a hot issue in Computer Science but also is a spotlight in the fields of Statistics and Sociology \cite{mehrabi2021survey, jobin2019global}, various definitions, testing techniques, and evaluating metrics have been proposed regarding to the fairness of machine learning algorithms and software \cite{zhang2020machine, zhang2021ignorance}. 
In this paper, we focus on the {\bf group fairness}, which is widely studied in Software Engineering,
and evaluate our approach with the measurements and baselines from the Software Engineering community \cite{sun2022causality, chakraborty2020fairway, chakraborty2021bias, chen2022maat, hort2021fairea}.

\subsection{Related Work}
As mentioned above, the rise of data-driven algorithms has made fairness issues widely studied. Some leading technology companies have also set up dedicated teams to explore fairer AI algorithms, applications, and software services. 
Where Microsoft created Fairlearn \cite{bird2020fairlearn},  an open-source toolbox that provides some general unfairness mitigation methods, and proposed an AI fairness list that determines the general requirements and focus of AI fairness \cite{madaio2020co}. 
Google AI came up with a series of studies \cite{hardt2016equality, beutel2019fairness, sambasivan2021everyone, lahoti2020fairness, prost2019toward} exploring the impact of the dataset, threshold, ranking, and re-weighting on AI fairness with various perspectives and methods. 

\textbf{Fairness testing and measure metrics}: 
As fairness problems receive high attention from many interdisciplinarities, how to test and evaluate model fairness is becoming an emergency issue. Zhang \etal \cite{zhang2020machine}  investigated 144 machine learning software fairness papers that cover attributes testing (fairness, robustness and correctness), testing components (dataset, algorithms and framework), testing workflow (test generation and evaluation) and application scenario (recommendation system, machine translation and autonomous driving) to explore fairness testing techniques.  

In terms of fairness testing, Galhotra \etal \cite{galhotra2017fairness}  provided a definition of software fairness and discrimination and proposed Themis, a method that can qualitatively test the fairness and bias degree; Udeshi \etal \cite{udeshi2018automated} proposed an automated directed fairness testing approach Aeqitas to test any given model’s fairness; Aggarwal \etal \cite{aggarwal2019black} came up with a black box fairness testing approach for a machine learning model via symbolic execution and local explainability for effective test case generation. 

As for fairness evaluation, Biswas \etal \cite{biswas2020machine} performed a comprehensive empirical study of the real-world machine learning model and built a baseline for 40 top Kaggle models in 5 tasks; Hort \etal \cite{hort2021fairea}  considered the problem of balancing model performance and fairness and came up with a performance-fairness trad-off baseline (Fairea Baseline). 

\textbf{Bias mitigation}: 
Zhang \etal \cite{zhang2021ignorance} found that extending the feature set significantly improves fairness, while enlarging training data does not improve fairness. According to different working mechanisms, existing bias mitigation approaches can be divided into three classes: pre-processing, in-processing, and post-processing \cite{mehrabi2021survey}, where pre-processing methods \cite{zemel2013learning, kamiran2012data, chakraborty2020fairway, chakraborty2021bias, peng2021xfair} take effects before decision-making algorithms via processing the dataset with data mutation or differentiated sampling; 
in-processing methods \cite{beutel2019fairness, zhang2018mitigating, rastegarpanah2019fighting, wu2021fairness, chen2022maat} optimize the decision-making algorithms via introducing the regular item, Constrained Markov Decision Process (CMDP), or group selection;
post-processing methods \cite{li2021user, fu2020fairness, tsintzou2018bias, kamiran2012decision} do not modify the data or decision-making algorithm but adjust the decision results with re-ranking, greedy algorithm, or bias disparity. IBM developed the open-source toolbox AIF360 \cite{bellamy2019ai}, which integrated fairness metrics and some bias mitigation algorithms, including pre-processing \cite{zemel2013learning, kamiran2012data}, in-processing \cite{zhang2018mitigating, kamishima2012fairness} and post-processing \cite{pleiss2017fairness, hardt2016equality, kamiran2012decision} methods.

Among various mitigation strategies, pre-processing methods are preferred in most cases \cite{biswas2020machine} because once the model is trained, users can utilize the discrimination-free model to make decisions directly instead of calling in/post-processing approaches in each decision-making task. Compared with in/post-processing methods, pre-processing methods make all complicated calculations and bias-mitigating operations before the trained model and output a normal but fair machine learning model, which does not increase the technical requirements and time or computation cost for the model users. As a result, pre-processing bias-mitigating methods can be easily deployed in real applications. That is why most machine learning fairness studies from Software Engineering are pre-processing solutions in recent years. 

For instance, Chakraborty \etal proposed Fairway \cite{chakraborty2020fairway}, which designs a pre-experiment to identify the biased data instances and remove them to improve fairness. As removing too many data instances can decrease model prediction performance, Chakraborty \etal introduced SMOTE \cite{fernandez2018smote} and proposed Fair-SMOTE \cite{chakraborty2021bias} in their following work. Fair-SMOTE solves Fairway's limitation by synthesizing new data instances to compensate for data removal. Furthermore, Chen \etal came up with an ensemble approach MAAT \cite{chen2022maat}, training the fairness model and performance model separately and improving fairness by combining two models to make the final decision. Existing bias-mitigating approaches can work as the fairness model of MAAT.

\subsection{Causality}
Causality (Causation) denotes the impact that one event, process, state, or object (a cause) contributes to the production of another event, process, state, or object (an effect) \cite{bunge2017causality}. At the same time, Causality can either refers to a science studying the impact mentioned above with a solid theoretical foundation or a technique processing the causation among variables \cite{pearl2018book}. 

Due to mature theories (Causal Intervention, Counterfactual Inference) and practical tools (Structural Causal Model, Causal Graph, Average Treatment Effect and Randomized Control Trials), Causality is widely used in Physiology, Biomedicine, and Economics, as there is a tremendous demand of processing data correlation in such areas \cite{pearl2009causality, gopnik2007causal, imbens2015causal, varian2016causal}. Likewise, as machine learning software encounters challenges in handling complicated correlations and bias among variables in datasets, Causality can be a potential direction to address such problems and obtain fairer machine learning software.

\subsubsection{Structural Causal Mode and Causal Graph}
\begin{figure}[!h]
\includegraphics[width=0.45\linewidth]{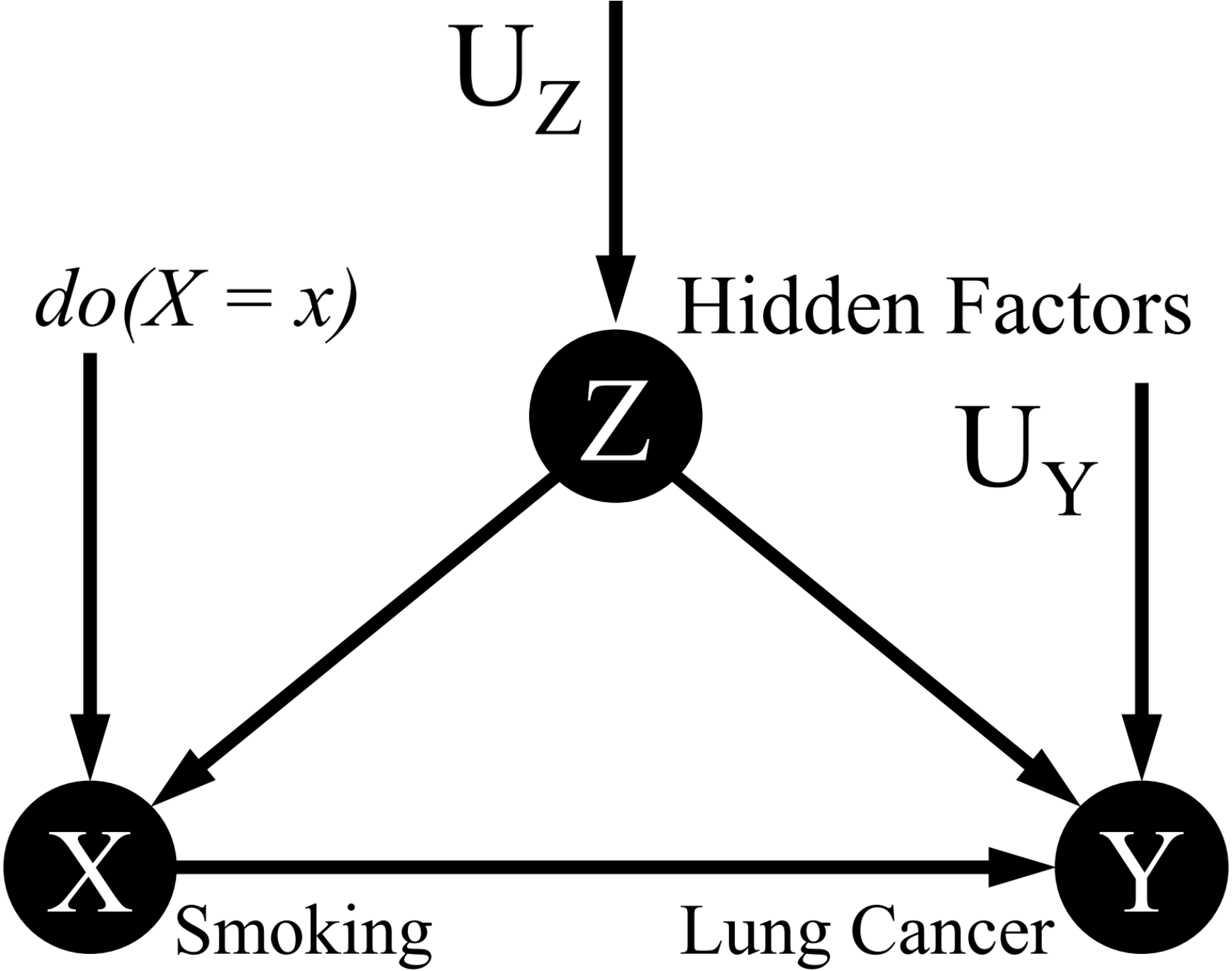}
\caption{Causal Graph of Smoking on Lung Cancer.}
\label{fig: Causal Graph of Smoking on Lung Cancer} 
\vspace{-2ex}
\end{figure}

Structural Causal Model (SCM) is a mathematical model representing the causation among features or variables, which can transform into a graph model (Causal Graph) in the form of a Directed Acyclic Graph (DAG) \cite{pearl2009causality}. As a  4-tuple model $M(X, U, f, P_{u})$, Structural Causal Model consists of endogenous variables $X$, exogenous variables $U$, mapping functions $f$, and $P_{u}$ which is the probability distribution over $U$ \cite{pearl2009causality, sun2022causality}. SCM and Causal Graph are essential tools of Causality Science to deal with causation. Figure \ref{fig: Causal Graph of Smoking on Lung Cancer} shows a causal graph of smoking on lung cancer.

\subsubsection{Causal Intervention}
Causal Intervention is a widely-used approach to calculate the potential effect of an untaken action or strategy, which is broadly applied in medical treatment, social affairs, and policy analysis \cite{pearl2009causality, heckman2000causal,kuipers1984causal}. In the following, We introduce Causal Intervention through the example of the Smoking-Lung Cancer case.

Suppose $P_{m}$ is the modified probability distribution (\ie the probability after intervention). According to the definition of intervention \cite{pearl2009causality}, there are:
\begin{equation}
P(Y = y|do( X = x)) = P_{m}(Y = y|X = x)
\end{equation}
\begin{equation}
P_{m}(Z = z) = P(Z = z)
\end{equation}
\begin{equation}
P_{m}(Y = y|Z = z, X = x) = P(Y = y|Z = z, X = x)
\end{equation}

where $Y$ denotes lung cancer, $X$ denotes smoking, $Z$ denotes hidden factor, $do( X = x)$ means taking intervention action of setting $X = x$.
Then we can obtain:

\begin{equation}
\begin{aligned}
\label{adjustment formula}
&P(Y = y|do( X = x)) = P_{m}(Y = y|X = x)\\
&= \sum_{z} P_{m}(Y = y|X = x, Z = z)P_{m}(Z = z|X = x)\\
&= \sum_{z} P_{m}(Y = y|X = x, Z = z)P_{m}(Z = z)\\
&= \sum_{z} P(Y = y|X = x, Z = z)P_{m}(Z = z)\\
\end{aligned}
\end{equation}
Eq (\ref{adjustment formula}) is the adjustment formula, and this operation is named \textit{adjusting for Z}, which facilitates us to use known probability distribution to calculate unknown intervened probability distribution \cite{pearl2009causality}. In the smoking-lung cancer case, it enables us to analyze the causal effect of smoking on lung cancer quantitatively.

%% file: 3.approach.tex
\section{APPROACH}
\label{sec:approach}
In this section, we introduce FITNESS in detail.

\begin{figure}[!t]
\includegraphics[width=0.97\linewidth]{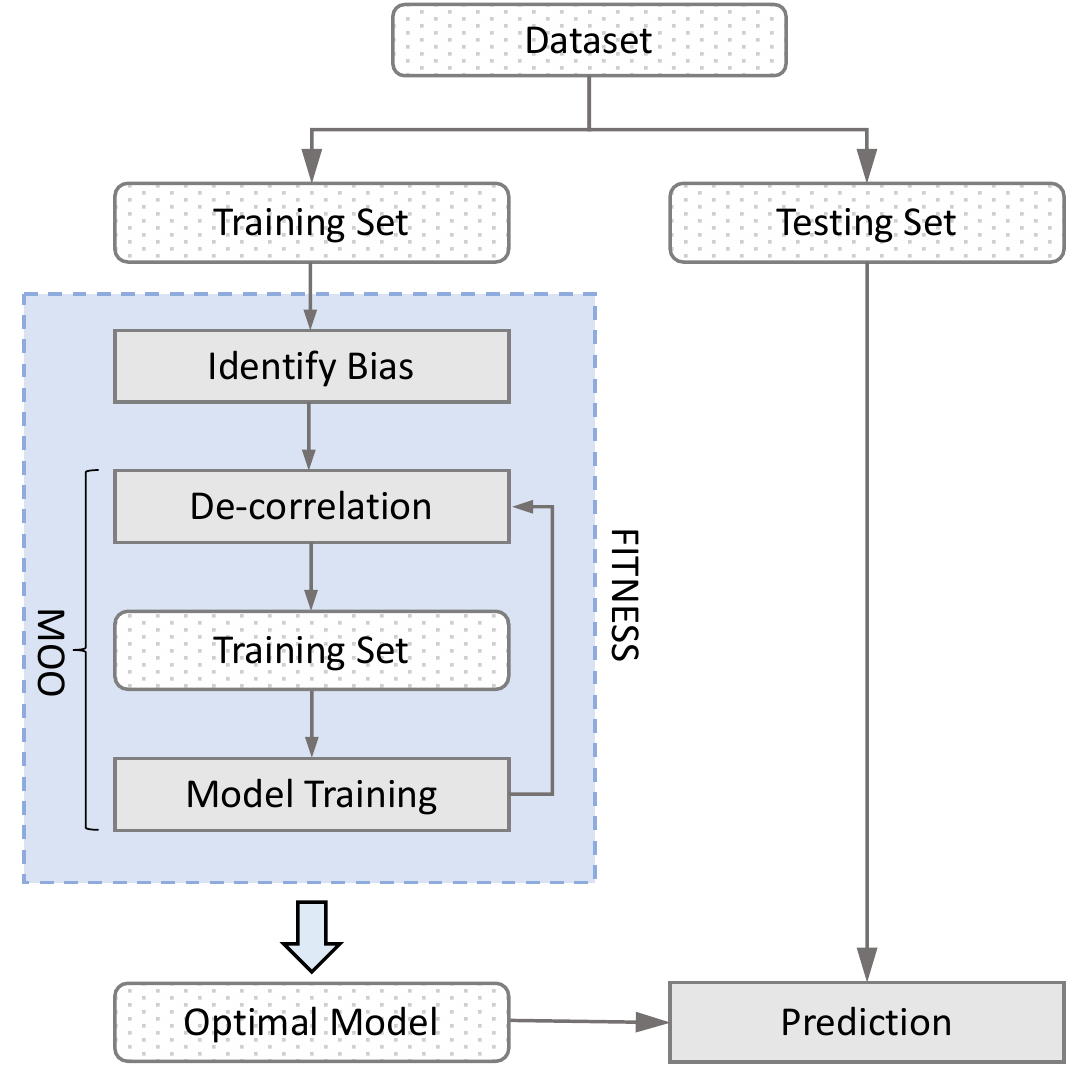}
\caption{Overview of FITNESS.}
\label{fig: overview} 
\vspace{-2ex} 
\end{figure} 

\subsection{Overview}
FITNESS is a pre-processing approach mitigating bias by causal de-correlation. Many existing works \cite{chakraborty2020fairway, chakraborty2021bias, hort2021fairea, zhang2021ignorance} mentioned mutating data instances, removing biased data instances and synthesizing data instances can modify the bias degree of data. However, what kinds of and how many data points should be mutated, removed or synthesized are still intractable challenges. FITNESS can automatically complete such processes and de-correlate the causal differences among various sensitive feature values on labels to achieve a fairness-performance trade-off.

FITNESS consists of two components, including bias identification and causal de-correlation. Figure \ref{fig: overview} shows the workflow of FITNESS. We first randomly divide the dataset into the training set and testing set. Then FITNESS conducts bias identification and causal de-correlation on the original training set by causality analysis to generate a new training set for the training model. Directed by the result of bias identification, the selected multi-objective optimization algorithm can nudge the optimal training set for training a fair and high-performance model by iterations.

\begin{figure}[!h]
\includegraphics[width=0.6\linewidth]{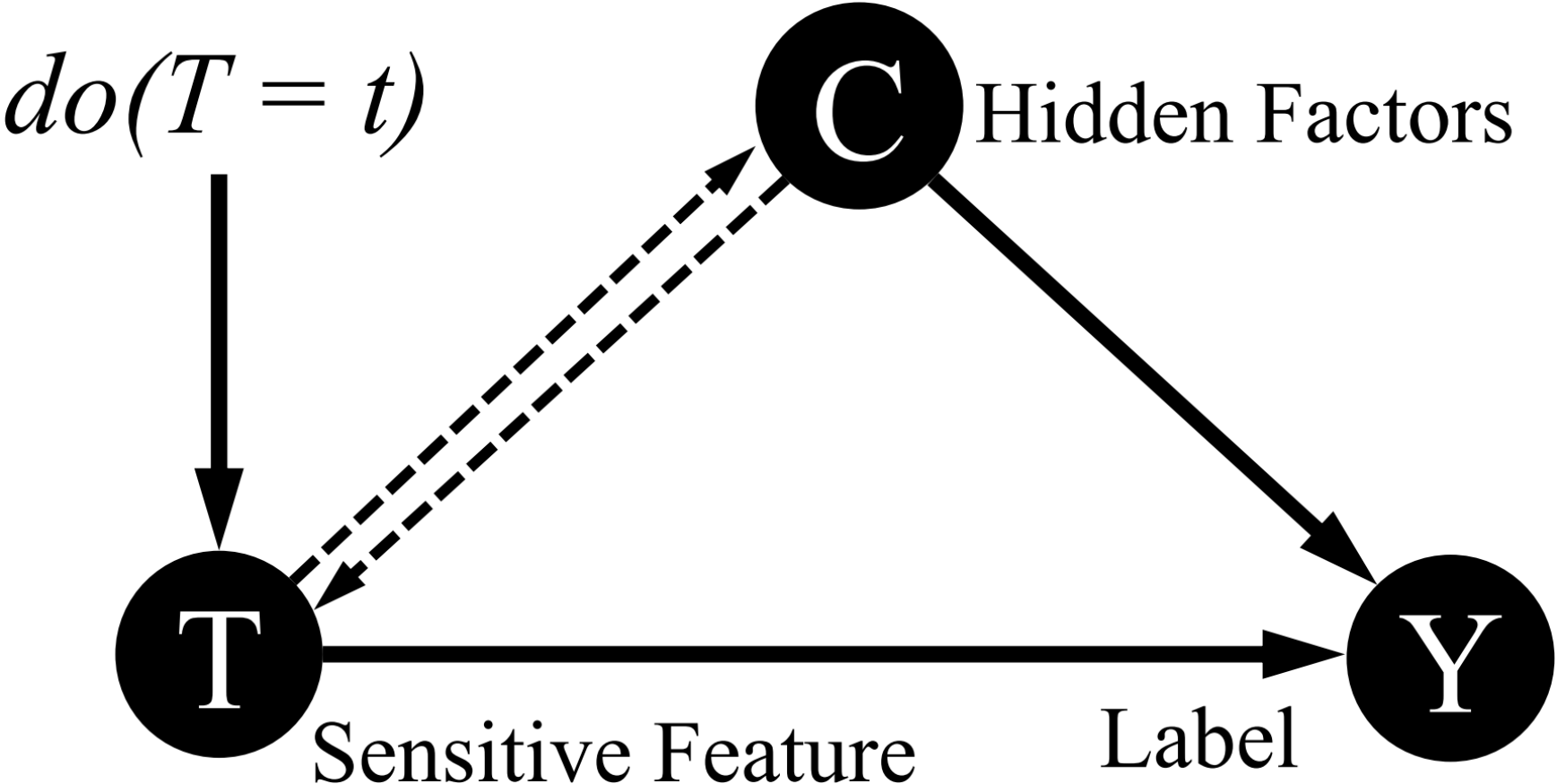}
\caption{Causal Graph of Sensitive Feature on Label.}
\label{fig: Causal Graph of Sensitive Feature on Label} 
\vspace{-3ex} 
\end{figure}

\subsection{Bias Identification}

Figure \ref{fig: Causal Graph of Sensitive Feature on Label} shows a causal graph of sensitive feature to label, where $T$ denotes the sensitive feature, $C$ denotes hidden factors (all other features and factors except the sensitive feature), and $Y$ represents the data label (\eg income < 50k in Adult Dataset). In Figure \ref{fig: Causal Graph of Sensitive Feature on Label}, both sensitive feature and hidden factors affect the label value causally. The label value $Y$ can be represented by a binary function:
\begin{equation}
Y = f(C, T) 
\label{eq: causal function(C, T)}
\end{equation}

Our goal is to make the causal effect of different sensitive feature values on the label equal. We expect that after causal de-correlation, the binary function Eq \ref{eq: causal function(C, T)} can be rewritten as $Y = f(C, T_c) $, where $T_c$ is a constant, the value of $Y$ is not affected by different sensitive feature values (\eg male or female). As a result, we can protect the sensitive feature and train discrimination-free models. To achieve this, we first identify the bias.

In most cases, it is convenient and effective to determine bias by comparing the percentage of feature and label classes. However, datasets may exist in Simpson's paradox \cite{pearl2009causality}, a phenomenon in which the relationship between two variables is reversed when they are analyzed in subgroup and overall, making it unreliable to identify bias by simply comparing the classes' percentages within the dataset. To avoid this case, we take causality analysis to identify the dataset's bias correctly.

For the convenience of causality analysis,
we define the effect of sensitive feature $T$ on label $Y$ within the dataset as the Average Causal Effect (ACE). According to literature \cite{pearl2009causality, sun2022causality}, ACE can be represented as:



\begin{equation}
ACE^{y_{i}}_{t_{j}}=\mathbb{E}[Y=y_{i}|do(T=t_{j})]
\label{eq: ACE}
\end{equation}
where $y_i$ denotes the $i$-th label (e.g., income > 50K in the Adult dataset), $t_j$ denotes the $j$-th sensitive feature value (e.g., female of sex in the Adult dataset), $do(T=t_{j})$ denotes taking causal intervention. $ACE^{y_{i}}_{t_{j}}$ means the average causal effect of $t_j$ against $y_i$. Through comparing $ACE^{y_{i}}_{t_{j}}$ and $ACE^{y_{i}}_{t_{k}}$, we can conveniently and correctly identify whether $t_{j}$ (\eg female in Adult Dataset) or $t_{k}$ (\eg male in Adult Dataset) is the privileged sensitive feature value on label $y_{i}$ (\eg income > 50K in Adult Dataset ).

\subsection{Causal De-correlation}
To automatically de-correlate sensitive feature values and data labels, we first define the Average Causal Difference (ACD)
to guide the searching algorithm to determine which data instances should be mutated. According to Eq \ref{eq: ACE}, ACD can be represented as:
\begin{equation}
ACD^{y_{i}}_{(t_{j},t_{k})}=ACE^{y_{i}}_{t_{j}}-ACE^{y_{i}}_{t_{k}}
\end{equation}
where $t_{j}$ and $t_{k}$ denote the different values (\eg female or male in Adult Dataset) of the same sensitive feature $T$ (\eg sex in Adult Dataset). If $ACD^{y_{i}}_{(t_{j},t_{k})} > 0$, the causal effect of $t_j$ is higher than $t_k$ on $y_i$, regarding $y_i$, $t_j$ will be the privileged sensitive feature value and the group with $t_k$ is likely to be disadvantaged. On the contrary, If $ACD^{y_{i}}_{(t_{j},t_{k})} < 0$, the causal effect of $t_k$ is higher than $t_j$ on $y_i$, regarding $y_i$, $t_k$ will be the privileged sensitive feature value and the group with $t_j$ is likely to be disadvantaged. An $ACD^{y_{i}}_{(t_{j},t_{k})} = 0$  implies the training dataset is unbiased regarding corresponding sensitive feature values, and this is exactly what we expect to achieve.

Many previous works \cite{pearl2009causality, chakraborty2020fairway, chakraborty2021bias, hort2021fairea} demonstrate that pre-processing operations (\eg reweighting, removing, synthesizing or mutating) can change the bias degree of a dataset. We utilize Average Causal Difference (ACD) to formulate this problem. For example, consider the following mutation.



\begin{equation}
Instance^{y_{i}}_{t_{j}}\longrightarrow Instance^{y_{l}}_{t_{k}} \nonumber
\end{equation}
where the operation means a data instance with sensitive feature value $t_j$ and label $y_i$ is mutated to the data instance with sensitive feature value $t_k$ and label $y_l$. According to the literature \cite{hort2021fairea}, this mutation can lead to unknown effects on data bias degree. We use the change of ACD as a metric to describe the change in data bias. So, the ACD in the new training set after data instances mutation can be written as:
\begin{equation}
\label{eq: ACD after mutation}
    ACD^{y_{i}}_{m(t_{j},t_{k})}=ACD^{y_{i}}_{(t_{j},t_{k})} + \Delta _{ACD}
\end{equation}
where $\Delta _{ACD}$ is the change of the original average causal difference caused by the mutation operation. Causal de-correlation is to make $ACD^{y_{i}}_{m(t_{j},t_{k})}$ approach to zero via regulating $\Delta _{ACD}$. To achieve causal de-correlation on the training dataset, we introduce a parameter $\alpha$ defined as follow:
\begin{equation}
    \alpha = \frac{n_{mutated}}{n_{totals}}
\end{equation}
where $n_{mutated}$ denotes the number of mutated instances and $n_{totals}$ denotes the totals of the original instances. When we make $ACD^{y_{i}}_{m(t_{j},t_{k})}$ approach to zero by regulating $\alpha$, we achieve the causal de-correlation operation on a single sensitive feature $T$. At this point, different values of a specific sensitive feature have the same effect on a particular label. The model trained from the causal de-correlated training set will fit the function $Y = f(C, T_c)$ where $T_c$ is a constant.

Regarding multiple sensitive features protection, we can introduce more mutation parameters to extend our approach, which would not lead to unbearable computation and time consumption.

\subsubsection{Multi-Objective Optimization}

According to the introduction above, we can find suitable parameters $\alpha$ (or combinations of $\alpha _1, \alpha _2, ..., \alpha _n$) that achieve causal de-correlation on the given training dataset. 
However, causal de-correlation involves mutating data instances where there is a risk of unbiased data instances being mutated. As a result, de-correlation may damage the original dataset and further affect the model prediction performance. To address this problem, we introduce Multi-Objective Optimisation (MOO) \cite{censor1977pareto}. The objective function, which combines fairness and performance metrics, allows the optimization algorithm to mutate appropriate data instances to achieve a trade-off between performance and fairness.

In this paper, we propose the following objective function $f(\alpha)$:

\begin{equation}
\begin{aligned}
\label{objective function}
f(\alpha) &= f_{1}(\alpha) + f_{2}(\alpha) + f_{3}(\alpha) + f_{4}(\alpha), \alpha \in [0, 1]
\end{aligned}
\end{equation}
where $f_{1}$ denotes the F1 Score metric, $f_{2}$ denotes the accuracy metric, $f_{3}$ denotes the EOD metric, $f_{4}$ denotes the AOD metric.
Specifically, the F1 score and accuracy are metrics on the performance of the model, while EOD and AOD are metrics on the fairness of the model (details of these metrics will be introduced in the following section). It is worth emphasizing that when necessary one can introduce more metrics or conditions to the multi-objective function.



\subsubsection{Particle Swarm Optimization (PSO)}
In our previous introduction, we used MOO to find optimal mutation parameters that balance fairness and model performance. It is worth noting that the feasible mutation parameters have a vast search space, and enumerating all possible solutions is computationally infeasible and inefficient. Therefore, we need to use more efficient search algorithms to find the optimal solution we need. Genetic Algorithms (GA) \cite{whitley1994genetic}, A* Search \cite{kohn1974theory}, Simulated Annealing (SA) \cite{van1987simulated}, and Particle Swarm Optimization (PSO) \cite{eberhart1995particle} are all excellent optimization algorithms. In this paper, we use PSO as the search algorithm for MOO because it has the characteristics of being simple, efficient, and easy to implement. PSO, which was invented by Dr Eberhart and Dr Kennedy is a random searching algorithm based on population cooperation. It simulates the foraging behavior of bird flocks and adjusts the speed and granularity of the search by setting parameters such as population size, the number of iterations, individual factors, and group factors. Upon convergence to the optimal solution that satisfies the exit criteria, PSO terminates the search process. 


%% file: 4.evaluation.tex
\section{EVALUATION}
\label{sec:exp}
In this part, we mainly evaluate the performance and applicability of FITNESS. We first present the benchmark datasets and evaluation metrics and then give the detailed experiment results and analyses. All the experiments were conducted on a Linux Server with environments of Ubuntu 20.04 focal, 256 GB RAM, 3.7 GHz Intel Xeon Gold 6238, and Python 3.7.



\subsection{Benchmark Datasets}
\label{sec:experiment:dataset}
To evaluate FITNESS, we select four widely-used datasets, including the Adult Census Income \cite{misc_adult_2}, Compas \cite{misc_compas_dataset}, German Credit \cite{misc_south_german_credit_522}, and Bank Marketing \cite{misc_bank_marketing_222}, to conduct the experiment. All the four datasets are class and sensitive features imbalanced, which makes decision-making systems trained on them face the problem of fairness.
The information of the four datasets is presented in Table~\ref{table 1}~\cite{chakraborty2021bias}.
Following previous studies~\cite{chen2022maat, chakraborty2021bias}, we introduce the following terminologies   to better analyze the effects of the biased data. 
Given a label, if it benefits or indicates an advantage for an instance, we call it a \textit{\textbf{favorable label}} (e.g., high income). Otherwise, we call it an \textit{\textbf{unfavorable label}} (e.g., low income).
{According to the sensitive feature, data samples can be divided into two groups. A group is called a \textbf{\textit{privileged group}} if the model prefers to give it a favorable label. A group is called an \textbf{\textit{unprivileged group}} if the model prefers to give it an unfavorable label.}
If the privileged group and unprivileged group are treated equally or similarly by the model, then the model is deemed to meet the \textit{\textbf{group fairness}} requirement.

\input{exp_dataset}

\subsection{Evaluation Metrics}
In this paper, we use four metrics (i.e.,  precision ($\frac{TP}{TP + FP}$), recall ($\frac{TP}{TP + FN}$) and F1 Score ($\frac{2*Precision*Recall}{Precision + Recall}$), Accuracy ($\frac{TP + TN}{TP+FP+TN+FN}$)) to evaluate model performance \cite{chakraborty2020fairway, chakraborty2021bias, chen2022maat}.
In this part, we denote TP as \textit{True Positives}, TN as \textit{True Negatives}, FP as \textit{False Negatives}, FN as \textit{False Negatives}, $P$ as Precision, $R$ as Recall, $Acc$ as Accuracy.

For convenience, we encode the terminologies used in measuring model fairness and evaluating our method. Supposing a given sensitive feature $\textbf{A}$, we encode the privileged group to $\textbf{1}$, encode the unprivileged group to $\textbf{0}$ as well as encode the favorable label to $\textbf{1}$, unfavorable label to $\textbf{0}$. Then according to the literature \cite{chen2022maat, chakraborty2021bias}, the fairness metrics can be calculated as follow, where $\hat{Y}$ demotes predicted label and $Y$ denotes real label.

\begin{itemize}[leftmargin = 10 pt]
    \item \textbf{EOD} (Equal Opportunity Difference) indicates the TP rate difference between privileged and unprivileged groups.
\end{itemize}
\begin{equation}
   EOD = P[\hat{Y}=1|A=0,Y=1]-P[\hat{Y}=1|A=1,Y=1]
\end{equation}

\begin{itemize}[leftmargin = 10 pt]
    \item \textbf{AOD} (Average Odds Difference) indicates the average of the FP rate difference and the TP rate between privileged and unprivileged groups.
\end{itemize}

\begin{equation}
\begin{aligned}
   AOD = &0.5[(P[\hat{Y}=1|A=0,Y=0]-P[\hat{Y}=1|A=1,Y=0])\\
   &+(P[\hat{Y}=1|A=0,Y=1]-P[\hat{Y}=1|A=1,Y=1])] 
\end{aligned}
\end{equation}

\begin{itemize}[leftmargin = 10 pt]
    \item \textbf{SPD} (Statistical Parity Difference)indicates the difference of probabilities of favorable label gained by privileged and unprivileged groups.
\end{itemize}
\begin{equation}
   SPD = P[\hat{Y}=1|A=0]-P[\hat{Y}=1|A=1]
\end{equation}

\begin{itemize}[leftmargin = 10 pt]
    \item \textbf{Fairness and Performance Trade-off Baseline.} To measure the effectiveness of the trade-off of the model fairness and performance,  at ESEC/FSE 2021, Hort \etal proposed Faireal \cite{hort2021fairea}, a trade-off baseline using the AOD-Accuracy and SPD-Accuracy metrics. To make the baseline more comprehensive, Chen \etal proposed MAAT \cite{chen2022maat}, extending the evaluation to fifteen fairness-performance metrics (combinations of three fairness metrics and five performance metrics) at ESEC/FSE 2022. The baseline divides the fairness-performance trade-off effectiveness into five levels, including "win-win" (improve both performance and fairness), "good" (improve fairness, reduce performance but surpass the Fairea Baseline ), "inverted" (improve performance but reduce fairness), "poor" (improve fairness, reduce performance but not surpass the Fairea Baseline) and "lose-lose" (reduce both fairness and performance). In this work, we adopted the extended Fairea Baseline from MAAT to evaluate our approach.
\end{itemize}



\subsection{Existing Approaches}
We widely take the state-of-the-art bias mitigating methods from the ML community and SE community for comparison. The baselines include pre-processing, in-processing, and post-processing methods. Where REW \cite{kamiran2012data}, ADV \cite{zhang2018mitigating}, and ROC \cite{kamiran2012decision} come from ML communities and were integrated into the IBM AIF360 toolkit. Fairway \cite{chakraborty2020fairway}, Fair-SMOTE \cite{chakraborty2021bias} and MAAT \cite{chen2022maat} come from reputable SE venue ESEF/FSE, and xFAIR \cite{peng2021xfair} comes from reputable SE journal TSE.

\subsubsection{Pre-processing Bias-Mitigating Methods}

\begin{itemize}[leftmargin = 10 pt]
    \item \textbf{REW \cite{kamiran2012data}}: REW set different weights to varying sensitive feature types to balance the weighted distribution of groups for mitigating dataset bias.
    \item \textbf{Fairway \cite{chakraborty2020fairway}}: Fairway designs a pre-experiment to identify the biased data points and remove them to obtain an unbiased training set for fairer ML models. 
    \item \textbf{Fair-SMOTE \cite{biswas2020machine}}: Fair-SMOTE synthesize some new data points to re-balance the training set class distribution.
    \item \textbf{xFAIR \cite{peng2021xfair}}: xFAIR mitigates model bias by replacing the original sensitive feature vector of the testing set with a synthesized sensitive feature vector.
\end{itemize}

\subsubsection{In/Post-processing Bias-Mitigating Methods}
\begin{itemize}[leftmargin = 10 pt]
    \item \textbf{ADV \cite{zhang2018mitigating}}: ADV introduces adversarial learning that hides the sensitive features and label in the training and validation process and simultaneously predicts the label and sensitive feature values simultaneously. If the model predicts the sensitive features value correctly, the model will be punished.
    \item \textbf{MAAT \cite{chen2022maat}}: MAAT is an ensemble approach training the fairness and performance models separately. It generates the final decision by combining the performance and fairness models with varying weights and achieves a good trade-off between model fairness and performance.
    \item \textbf{ROC \cite{kamiran2012decision}}: ROC's core idea is reversing the predicated result when the model gives a favorable label to a privileged group or gives an unfavorable label to an unprivileged group. 
\end{itemize}

\subsection{Experiment Settings}
In order to comprehensively evaluate FITNESS and ensure the convenience of reproducing our approach, we introduce our experimental settings in detail.

At first, we choose the four widely used benchmark datasets mentioned above. These four datasets have been integrated into IBM AIF360 \cite{bellamy2019ai} and broadly utilized in much bias-mitigating research \cite{kamiran2012data, kamiran2012decision, kamishima2012fairness, zhang2018mitigating, chakraborty2020fairway, chakraborty2021bias,chen2022maat, peng2021xfair}, which guarantees the reliability of our results. In addition, to evaluate the applicability in different machine learning algorithms, we followed the previous research \cite{chakraborty2020fairway, chakraborty2021bias, chen2022maat}, and conduct experiments in three typical ML algorithms including Logistic Regression (LR)~\cite{wright1995logistic}, Support Vector Machine (SVM) \cite{noble2006support} and Random Forest (RF) \cite{biau2016random}. To eliminate the effect of random confounds on the experimental results, we adopt the default configurations from the scikit-learn \cite{scikit-learn} library to realize each ML algorithm. Similarly, we implement the PSO algorithm (population size = 10, maximum iteration = 5) from the scikit-opt \cite{misc_Scikit-opt} library.

As for the implementation of baseline approaches, we call the library of IBM AIF360 to realize the bias-mitigating methods of REW, ADV and ROC. For the implementation of Fairway, Fait-SMOTE and MAAT, we use the source codes released by the paper authors or the codes
checked by the first authors \cite{chakraborty2020fairway, chakraborty2021bias, chen2022maat}. 
In each task, the datasets will split into a 70\% training set and a 30\% testing set, and each task will repeat 50 times and pick the means as the final result.

\subsection{Research Questions}
The effectiveness of FITNESS was evaluated via answering the following research questions.

\begin{itemize}[leftmargin = 10 pt]
   \item \textbf{RQ1(Effectiveness of improving fairness): } \textit{How significant can FITNESS achieve in improving fairness?} We compare FITNESS with existing methods in four performance metrics and three fairness metrics to demonstrate the bias-mitigating effectiveness.

    \item \textbf{RQ2(Balancing model performance and fairness): } \textit{What fairness-performance trade-off does FITNESS achieve?} We compare FITNESS with existing methods by Fairea Baseline to indicate the ability to achieve an excellent performance-fairness trade-off.
    
    \item \textbf{RQ3(Multiple sensitive features protection): } \textit{Can FITNESS protect multiple sensitive features simultaneously?}In a real-world application, the dataset may contain more than one sensitive feature need to be protected. This RQ explores whether FITNESS can protect multiple sensitive features as effectively as a single sensitive feature.
    
    \item \textbf{RQ4(Impact of multi-objective optimization strategies): } \textit{What is the impact of varying MOO strategies on the effectiveness of FITNESS?} In this research question, we explore the impact of different MOO strategies on FITNESS effectiveness by regulating the performance-fairness weight ratio.
    
\end{itemize}

%% file: exp_dataset.tex
\begin{table*}[t]
  \centering
  \caption{Benchmark datasets.}
  \vspace{-2ex}
  \resizebox{\textwidth}{!}{
    \begin{tabular}{|c|c|c|c|c|}%
        \hline
        \rowcolor{lightgray}
        \textbf{Dataset} & \textbf{Features} & \makecell{\textbf{Sensitive}\\ \textbf{Features}} & \textbf{Size} & \textbf{Description} \\
        \hline
        Adult Census Income & 14 & Sex, Race & 45,222 &  Individual information from 1994 U.S. census. Goal is predicting income >\$50,000.\\
        \hline
        Compas	& 28	& Sex, Race	& 6,167	& Contains criminal history of defendants. Goal is predicting re-offending in future.\\
        \hline
        German Credit & 20	& Sex	& 1,000	& Personal information about individuals \& predicts good or bad credit.\\
        \hline
        Bank Marketing	& 16	& Age	& 30,488	& Contains marketing data of a Portuguese bank. Goal is predicting term deposit.\\
        \hline
    \end{tabular}
    }
  \label{table 1}%
\end{table*}%

%% file: 5.results.tex
\section{Results}
\label{sec:eval}
In this section, we present and analyze the experiment results via answering the research questions. 

\subsection{RQ1: Effectiveness of Improving Fairness}
This research question evaluates whether FITNESS can significantly improve model fairness in various scenarios.
We design four experimental setups to compare the existing state-of-the-art approaches in different datasets and machine learning algorithms. The measurement consists of four performance metrics (accuracy, recall, precision, F1 score) and three fairness metrics (AOD, EOD, SPD). The setups include: (1) Compare with state-of-the-art pre-processing methods in different datasets and tasks; (2) Compare with state-of-the-art in/post-processing methods in different datasets and tasks; (3) Compare with state-of-the-art pre-processing methods in different machine learning algorithms; (4) Compare with state-of-the-art pre-processing methods in different machine learning algorithms.
Limited by space, we hand out the result of comparing FITNESS with pre-processing methods and in/post-processing methods in different datasets and tasks to display more results of our approach on this paper. Plus, as mentioned in \S \ref{sec:intro}, we also release full experimental results.

\input{exp_RQ1_1}
\input{exp_RQ1_2}

     
    
    
    

\subsubsection{\textbf{Results of different benchmark tasks.}}
According to Table \ref{tab: exp_RQ1_1}, we can obtain a general awareness that all FITNESS cells are located in the "good rank" area, and most FITNESS models realized the "best" value of corresponding metrics. In terms of REW and Fair SMOTE, although they can keep the model performance, their effectiveness in improving model fairness is not evident. On the contrary, Fairway and xFAIR improved model fairness but lost too much prediction performance in some tasks (Fairway in Adult-Sex and German-Sex tasks; xFAIR in Adult-Sex and Adult-Race tasks). Table \ref{tab: exp_RQ1_2} displays the result of comparison with in/post-processing methods. FITNESS keeps excellent effectiveness in improving model fairness among all datasets and tasks. As for the existing in/post-processing methods, ROC can mitigate model bias in the Adult dataset, and MAAT can improve model fairness in the Bank-Age task. ADV can not reduce any discrimination among the three tasks.

\subsubsection{\textbf{Results of different machine learning algorithms.}}
According to Table \ref{tab: exp_RQ1_3} and Table \ref{tab: exp_RQ1_4}, all bias-mitigating approaches achieve similar effectiveness in guaranteeing model prediction performance while they perform enormously vary in improving model fairness. In the Compas-Sex task, REW significantly reduces model bias in LR and SVM algorithms but fails in the RF. Fair SMOTE shows stable and good bias-mitigating effectiveness, but FITNESS outperforms all existing methods. As for the in/post-processing methods in Adult-Sex tasks, MAAT can significantly improve model fairness in the three machine learning algorithms, while FITNESS is still the best.

\input{exp_RQ1_3}
\input{exp_RQ1_4}

\noindent\textbf{Answer to RQ1:} Among all the experimental cases, FITNESS significantly improves the model fairness without losing too much model performance. The effectiveness is better than all the compared existing methods. Thus, we can confidently claim that FITNESS can significantly improve model fairness in various scenarios.

\subsection{RQ2: Balancing Performance and Fairness}
This research question explores whether FITNESS outperforms all the state-of-the-art approaches in trade-off model performance and fairness. In RQ1, we compared FITNESS with existing approaches in the effectiveness of mitigating bias. According to the results of RQ1, we can have a general awareness that FITNESS can significantly improve model fairness without losing too much model performance. However, Table \ref{tab: exp_RQ1_1}-\ref{tab: exp_RQ1_4} can not evaluate how effective the bias-mitigating approaches are in realizing the performance and fairness trade-off. Plus, the tables show the means of fifty run results, which may erase many details of a single run. 

To comprehensively evaluate the effectiveness of FITNESS and existing methods in balancing model performance and fairness, we introduce Faireal Baseline and compare with 6,300 cases (7 approaches $\times$ 6 tasks $\times$ 3 ML algorithms $\times$ 50 runs) in 15 metrics (5 performance metrics $\times$ 3 fairness metrics).

In terms of beating the Faireal Baseline, MAAT is the state-of-the-art method, which realizes 34.59\% "win-win" and 56.69\% "good" trade-off and beats Faireal Baseline in 91.28\% cases. FITNESS outperforms MAAT with 34.90\% "win-win" and 61.82\% "good" trade-off and beats Faireal Baseline in 96.72\% cases. In addition, we adopt the MAAT strategy introducing non-parametric Mann Whitney U-test \cite{mann1947test} to analyze the impact on model fairness and performance by the bias-mitigating approaches. As for each method, we compare 50 original models with 50 models that applied the bias-mitigating method and calculate the proportions of scenarios where it increases fairness and decreases performance.

\begin{figure}[!h]
\centering 
\setlength{\abovecaptionskip}{0.2cm}
\includegraphics[width=1\linewidth]{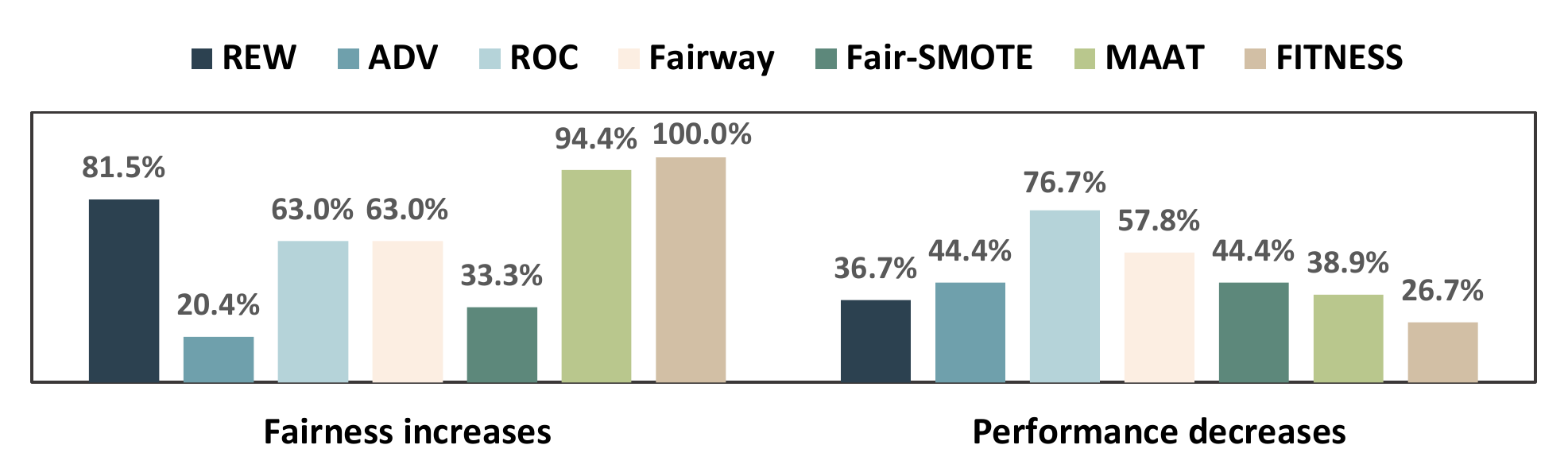}
\vspace{-2ex}
\caption{(RQ2) Proportion of scenarios where each approach significantly improves fairness and decreases performance.}
\label{fig: trade-off} 
\vspace{-1ex} 
\end{figure}

Figure \ref{fig: trade-off}  shows the result of the non-parametric Mann Whitney U-test among six existing methods and FITNESS. FITNESS increases the model fairness among 100\% scenarios (task and algorithms combinations), which beats the state-of-the-art method MAAT (increases model fairness among 94,44\% scenarios). Only 26.67\% models that applied FITNESS decreased model performance, significantly outperforming the state-of-the-art methods REW (36.67\%) and MAAT (38.89\%).

\begin{figure}[!h]
\centering 
\setlength{\abovecaptionskip}{0.2cm}
\includegraphics[width=1.\linewidth]{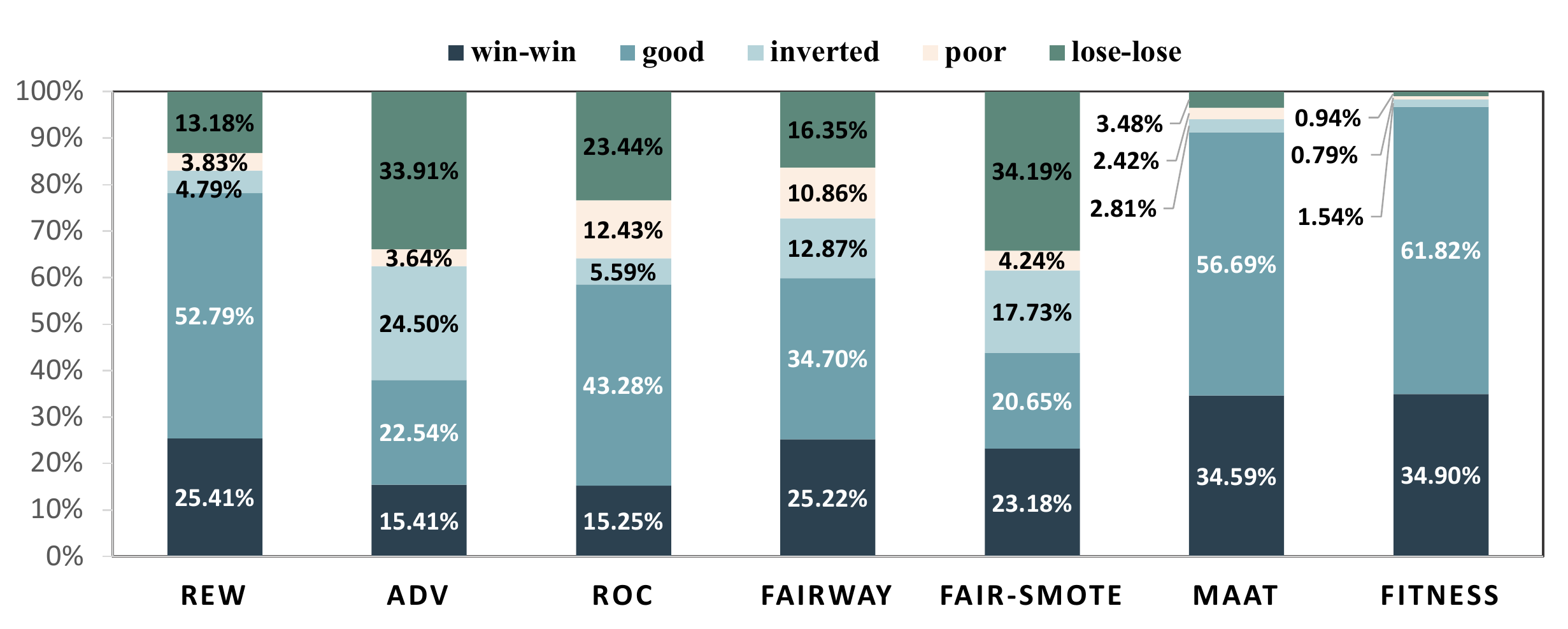}
\vspace{-2ex}
\caption{(RQ2) Effectiveness level distribution of FITNESS and existing approaches in benchmark tasks.}
\label{fig: Faireal_Baseline} 
\vspace{-1ex} 
\end{figure}

A detail that should be noticed is the result from Figure \ref{fig: Faireal_Baseline} and Figure \ref{fig: trade-off} seems to be conflict. The Mann Whitney U-test presents FITNESS increased model fairness among all scenarios, but the Faireal Baseline shows a bit of FITNESS models located in the "inverted," "poor," or "lose-lose" domain, which means some models that applied FITNESS decrease the fairness. This is because Mann Whitney U-test considers three fairness metrics (SPD, AOD, EOD) integrally while Faireal Baseline considers each fairness metric separately. The truth of the false conflict is that certain fairness metrics of a specific case that applied FITNESS decreased, but comprehensively considering three fairness metrics, the model fairness of the case still increased.

\noindent\textbf{Answer to RQ2:} FITNESS beats the Faireal Baseline in 96.72\% of all cases, which is higher than the best existing technique. Plus, FITNESS improves model fairness in 100\% cases and only decreases model performance in 26.67\% cases. It is FITNESS that improves more cases fairness and decreases fewer cases performance. Thus, we can answer this research question that FITNESS outperforms the state-of-the-art techniques in trade-off model fairness and performance.


\subsection{RQ3: Multiple Sensitive Features Protection}
We notice that a dataset may contain more than one sensitive feature that needs to be protected. This research question evaluates whether FITNESS can protect multiple sensitive features simultaneously.  In our benchmark datasets, the Adult and Compas datasets both contain "sex" and "race" that need to be protected. To our best knowledge, Fair-SMOTE, MAAT and xFAIR are state-of-the-art bias-mitigating approaches that can protect multiple sensitive features at the same time. Thus, we compare FITNESS with these three approaches to evaluate the effectiveness of protecting multiple sensitive features.

Table \ref{tab: exp_RQ3} presents the result of multiple sensitive features protection. In both datasets, FITNESS effectively maintains high model performance and improves model fairness. Almost all metrics values of the model that applied FITNESS are located in the top ranks. MAAT realized excellent bias-mitigating effectiveness in the Adult dataset but performed worse in the Compas dataset. xFAIR can achieve the best value on specific metrics (\eg AOD) but the worse on other metrics (\eg EOD in Adult-Sex and Compas-Race tasks). In addition, the prediction performance of the model that applied xFAIR fluctuates in different datasets, which reveals the instability of xFAIR.

\input{exp_RQ3}

\noindent\textbf{Answer to RQ3:} Among all the experimental cases, FITNESS significantly improves the model fairness without losing too much model performance. The effectiveness and stability are better than all the existing methods. Thus, we can claim that FITNESS can simultaneously protect multiple sensitive features.


\subsection{RQ4: Impact of Multi-Objective Optimization Strategies}

In this research question, we explore the impact of different performance and fairness weights on the effectiveness of FITNESS in improving model fairness and keeping performance. We conduct the experiments with a total of 5,400 cases (6 MOO strategies $\times$ 6 benchmark tasks $\times$ 3 ML algorithms $\times$ 50 runs). 

\begin{figure}[!h]
\centering 
\setlength{\abovecaptionskip}{0.1cm}
\subfigbottomskip=1pt
\subfigure[Proportion of scenarios where each MOO strategy significantly improves fairness and decreases performance.]{\includegraphics[width=1\linewidth]{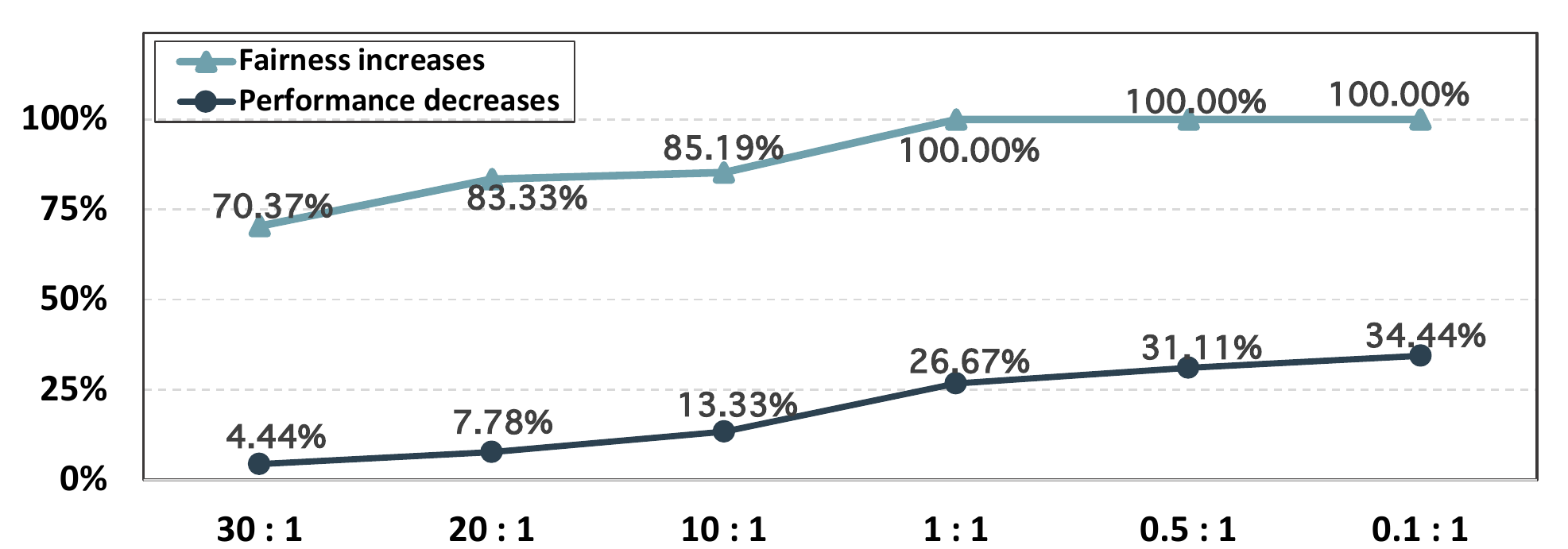} \label{fig: RQ4-1}}
\subfigure[Effectiveness level distribution of FITNESS with each MOO strategy.]{\includegraphics[width=1\linewidth]{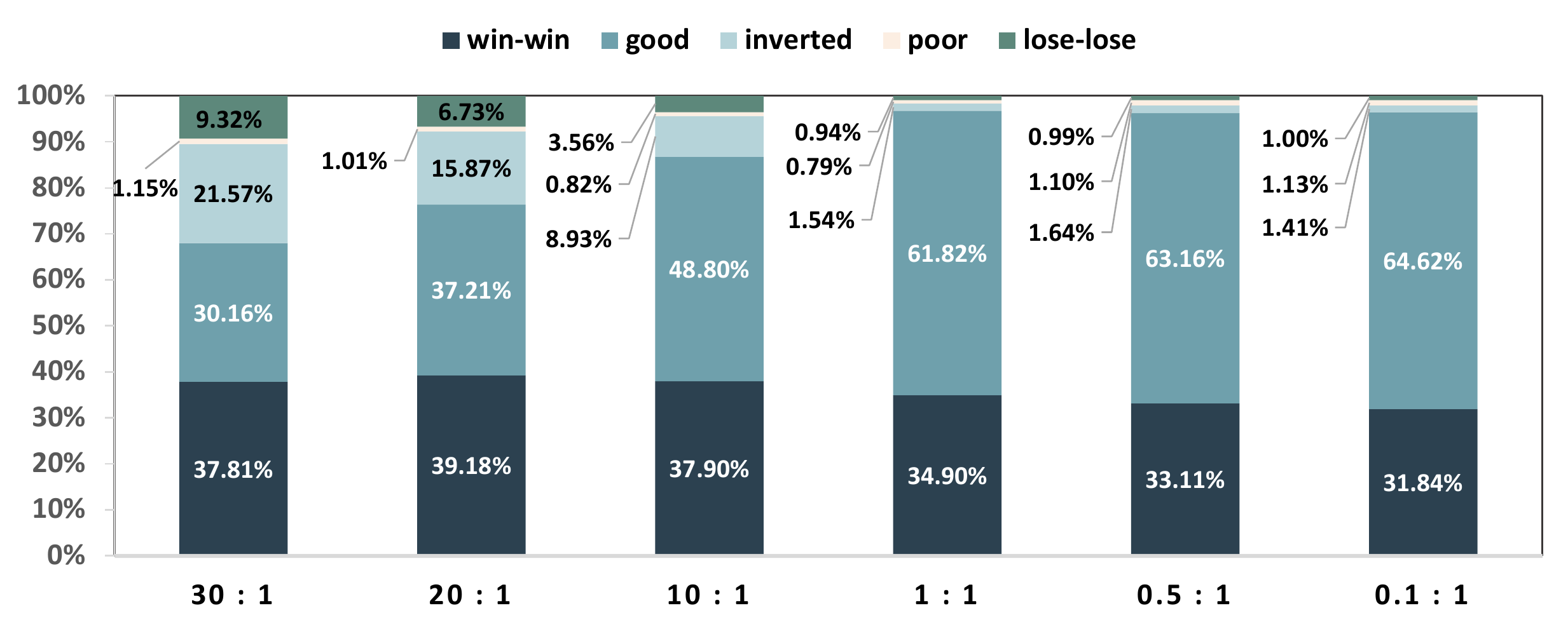}\label{fig: RQ4-2}}
\caption{(RQ4) Impact of different MOO strategies (weight ratio of performance and fairness from 30:1 to 0.1:1).}
\label{fig: RQ4-1-2 Impact of different MOO trategies}
\vspace{-3ex} 
\end{figure}

Figure \ref{fig: RQ4-1} presents the result that with the increasing of fairness weight, FITNESS can significantly improve the proportion of models whose fairness is enhanced from 70.37\% to 100.00\%. In terms of model performance, when decreasing the performance weight, the proportion of models whose loss prediction performance increases from 4.44\% to 34.44\%. In addition, when the proportion of the model whose fairness improved reaches 100.00\%, FITNESS can not improve the proportion via increased fairness weight. Continuing to increase the fairness weight can decrease more model prediction performance. Figure \ref{fig: RQ4-2} shows although a MOO strategy with a much higher performance weight can make more models located in the "win-win" domain, a similar weight for model performance and fairness can achieve better effectiveness in beating Faireal Baseline.

\begin{figure}[!h]
\centering 
\setlength{\abovecaptionskip}{0.1cm}
\includegraphics[width=1\linewidth]{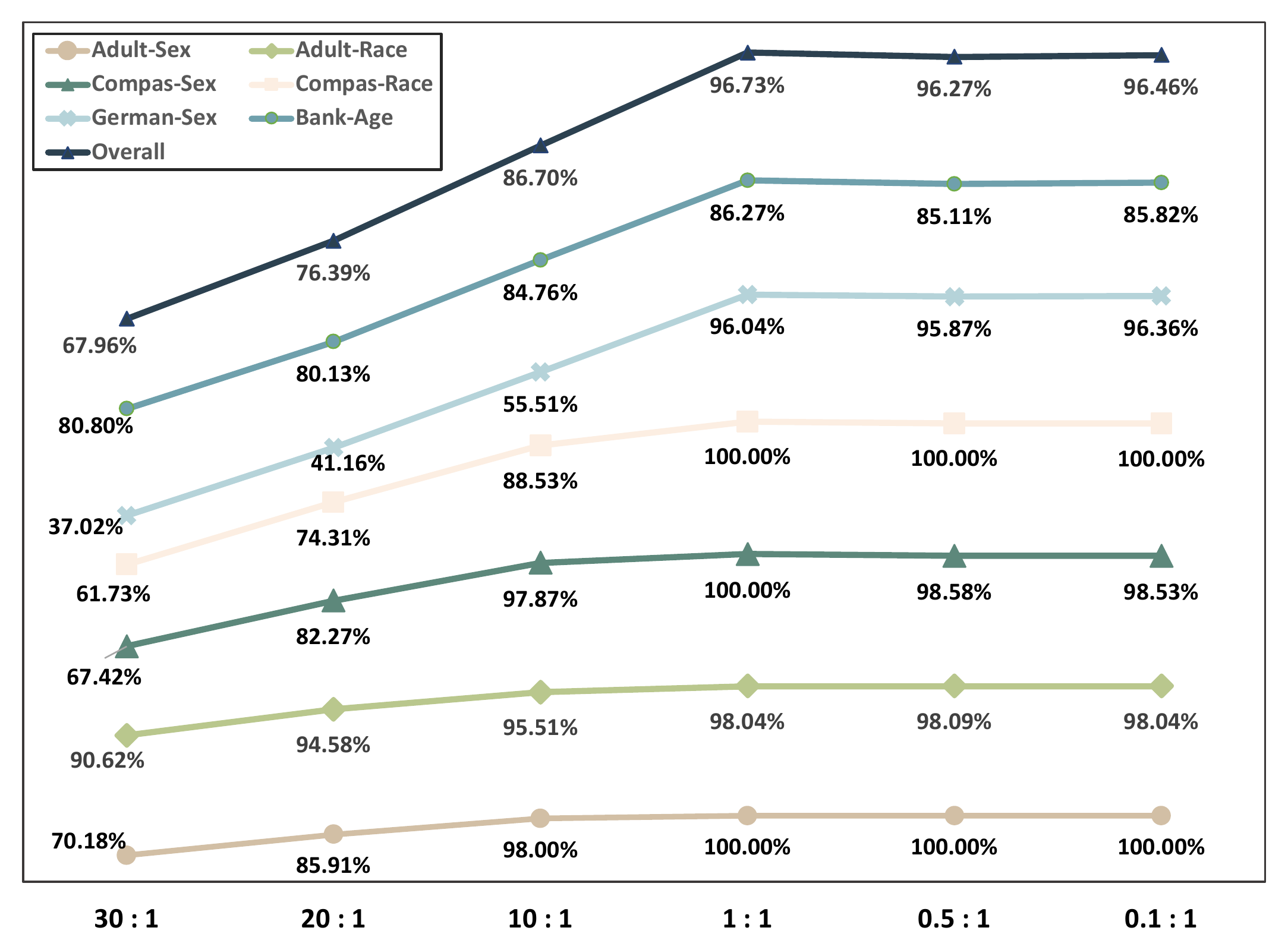}
\caption{(RQ4) Proportion of FITNESS surpassing Fairea Baseline with different MOO strategies (weight ratio of performance and fairness from 30:1 to 0.1:1) in each benchmark and overall task.}
\label{fig: RQ4-3}
\vspace{-3ex} 
\end{figure}

Figure \ref{fig: RQ4-3} presents that with the increases of fairness weight in the MOO strategy, the proportion of models beating the Faireal Baseline can increase overall, but the proportion will reach the peak near the similar performance-fairness weight strategy. Continuing to increase fairness can not increase the proportion anymore. As for each benchmark task, the optimal performance-fairness weight seems to vary.




\noindent\textbf{Answer to RQ4:} The weight of fairness and performance can significantly affect the effectiveness of balancing model fairness and performance. A higher fairness weight can realize the better
effectiveness in improving fairness, but the optimal weight balancing model fairness and performance is more likely near the equal performance-fairness weight. Plus, the optimal weight for each benchmark task needs further exploration as they typically vary.


%% file: exp_RQ1_1.tex
\begin{table}[h]
  \centering
    \caption{(RQ1) Results of comparing FITNESS with existing pre-processing methods in different tasks. "Default" means without applying any bias-mitigating operation; "Adult-Sex" means protecting sex feature in Adult dataset; "R" means recall; "P" means precision; "Acc" means accuracy; "F1" means F1 score and regarding these performance metrics, the higher, the better. As for fairness metrics AOD, EOD and SPD, the lower, the better. The \colorbox[rgb]{ .435,  .627,  .675}{darker} cells mean the top rank, the \colorbox[rgb]{ .71,  .827,  .851}{lighter} cells indicate the better rank, and the white cells mean worse rank. The value in each cell indicates the mean value of 50 runs. The same as Table \ref{tab: exp_RQ1_2} to Table \ref{tab: exp_RQ3}.
}
  \resizebox{\columnwidth}{!}{
        \begin{tabular}{|c|c|c|c|c|c|c|c|c|c|}
    \hline
    \rowcolor[rgb]{ .988,  .933,  .886} Alg.  & Task  & Method & Acc   & R     & P     & F1    & SPD   & AOD   & EOD \bigstrut\\
    \hline
    \multirow{18}[36]{*}{\begin{turn}{-90}Random Forest\end{turn}} & \multirow{6}[12]{*}{\begin{turn}{-90}Adult-Sex\end{turn}} & Default & \cellcolor[rgb]{ .435,  .627,  .675}0.84 & \cellcolor[rgb]{ .435,  .627,  .675}0.77 & 0.79  & \cellcolor[rgb]{ .435,  .627,  .675}0.78 & 0.19  & 0.08  & 0.08 \bigstrut\\
\cline{3-10}          &       & REW   & \cellcolor[rgb]{ .435,  .627,  .675}0.84 & \cellcolor[rgb]{ .71,  .827,  .851}0.76 & 0.79  & \cellcolor[rgb]{ .435,  .627,  .675}0.78 & 0.18  & 0.07  & 0.07 \bigstrut\\
\cline{3-10}          &       & Fairway & \cellcolor[rgb]{ .435,  .627,  .675}0.84 & 0.68  & \cellcolor[rgb]{ .435,  .627,  .675}0.85 & 0.72  & \cellcolor[rgb]{ .435,  .627,  .675}0.09 & \cellcolor[rgb]{ .71,  .827,  .851}0.03 & \cellcolor[rgb]{ .71,  .827,  .851}0.05 \bigstrut\\
\cline{3-10}          &       & Fair SMOTE & \cellcolor[rgb]{ .435,  .627,  .675}0.84 & \cellcolor[rgb]{ .435,  .627,  .675}0.77 & \cellcolor[rgb]{ .71,  .827,  .851}0.80 & \cellcolor[rgb]{ .435,  .627,  .675}0.78 & 0.19  & 0.08  & 0.07 \bigstrut\\
\cline{3-10}          &       & xFAIR & \cellcolor[rgb]{ .71,  .827,  .851}0.83 & 0.56  & \cellcolor[rgb]{ .71,  .827,  .851}0.67 & 0.61  & \cellcolor[rgb]{ .71,  .827,  .851}0.10 & \cellcolor[rgb]{ .435,  .627,  .675}0.02 & 0.06 \bigstrut\\
\cline{3-10}          &       & FITNESS & \cellcolor[rgb]{ .435,  .627,  .675}0.84 & \cellcolor[rgb]{ .71,  .827,  .851}0.73 & \cellcolor[rgb]{ .435,  .627,  .675}0.81 & \cellcolor[rgb]{ .71,  .827,  .851}0.76 & \cellcolor[rgb]{ .71,  .827,  .851}0.12 & \cellcolor[rgb]{ .435,  .627,  .675}0.02 & \cellcolor[rgb]{ .435,  .627,  .675}0.00 \bigstrut\\
\cline{2-10}          & \multirow{6}[12]{*}{\begin{turn}{-90}Adult-Race\end{turn}} & Default & \cellcolor[rgb]{ .71,  .827,  .851}0.84 & \cellcolor[rgb]{ .71,  .827,  .851}0.77 & \cellcolor[rgb]{ .71,  .827,  .851}0.79 & \cellcolor[rgb]{ .71,  .827,  .851}0.78 & 0.10  & 0.05  & 0.04 \bigstrut\\
\cline{3-10}          &       & REW   & \cellcolor[rgb]{ .71,  .827,  .851}0.84 & \cellcolor[rgb]{ .71,  .827,  .851}0.77 & \cellcolor[rgb]{ .71,  .827,  .851}0.79 & \cellcolor[rgb]{ .71,  .827,  .851}0.78 & 0.10  & 0.04  & 0.04 \bigstrut\\
\cline{3-10}          &       & Fairway & \cellcolor[rgb]{ .435,  .627,  .675}0.85 & \cellcolor[rgb]{ .71,  .827,  .851}0.76 & \cellcolor[rgb]{ .435,  .627,  .675}0.82 & \cellcolor[rgb]{ .435,  .627,  .675}0.79 & \cellcolor[rgb]{ .71,  .827,  .851}0.08 & 0.03  & 0.04 \bigstrut\\
\cline{3-10}          &       & Fair SMOTE & \cellcolor[rgb]{ .435,  .627,  .675}0.85 & \cellcolor[rgb]{ .435,  .627,  .675}0.78 & \cellcolor[rgb]{ .71,  .827,  .851}0.80 & \cellcolor[rgb]{ .435,  .627,  .675}0.79 & 0.10  & 0.04  & 0.04 \bigstrut\\
\cline{3-10}          &       & xFAIR & \cellcolor[rgb]{ .71,  .827,  .851}0.82 & 0.53  & 0.72  & 0.61  & \cellcolor[rgb]{ .435,  .627,  .675}0.07 & \cellcolor[rgb]{ .435,  .627,  .675}0.00 & \cellcolor[rgb]{ .71,  .827,  .851}0.03 \bigstrut\\
\cline{3-10}          &       & FITNESS & \cellcolor[rgb]{ .71,  .827,  .851}0.84 & \cellcolor[rgb]{ .71,  .827,  .851}0.77 & \cellcolor[rgb]{ .71,  .827,  .851}0.79 & \cellcolor[rgb]{ .71,  .827,  .851}0.78 & \cellcolor[rgb]{ .71,  .827,  .851}0.08 & \cellcolor[rgb]{ .71,  .827,  .851}0.02 & \cellcolor[rgb]{ .435,  .627,  .675}0.01 \bigstrut\\
\cline{2-10}          & \multirow{6}[12]{*}{\begin{turn}{-90}German-Sex\end{turn}} & Default & \cellcolor[rgb]{ .71,  .827,  .851}0.76 & \cellcolor[rgb]{ .71,  .827,  .851}0.66 & \cellcolor[rgb]{ .71,  .827,  .851}0.73 & \cellcolor[rgb]{ .71,  .827,  .851}0.67 & 0.07  & 0.07  & \cellcolor[rgb]{ .71,  .827,  .851}0.04 \bigstrut\\
\cline{3-10}          &       & REW   & \cellcolor[rgb]{ .71,  .827,  .851}0.75 & \cellcolor[rgb]{ .71,  .827,  .851}0.65 & \cellcolor[rgb]{ .71,  .827,  .851}0.72 & \cellcolor[rgb]{ .71,  .827,  .851}0.66 & 0.07  & 0.07  & \cellcolor[rgb]{ .71,  .827,  .851}0.04 \bigstrut\\
\cline{3-10}          &       & Fairway & 0.70  & \cellcolor[rgb]{ .71,  .827,  .851}0.64 & 0.67  & 0.63  & \cellcolor[rgb]{ .71,  .827,  .851}0.05 & 0.07  & 0.05 \bigstrut\\
\cline{3-10}          &       & Fair SMOTE & \cellcolor[rgb]{ .71,  .827,  .851}0.75 & \cellcolor[rgb]{ .71,  .827,  .851}0.68 & \cellcolor[rgb]{ .71,  .827,  .851}0.71 & \cellcolor[rgb]{ .71,  .827,  .851}0.69 & 0.06  & 0.06  & \cellcolor[rgb]{ .71,  .827,  .851}0.04 \bigstrut\\
\cline{3-10}          &       & xFAIR & 0.70  & \cellcolor[rgb]{ .435,  .627,  .675}0.92 & \cellcolor[rgb]{ .71,  .827,  .851}0.73 & \cellcolor[rgb]{ .435,  .627,  .675}0.81 & 0.08  & \cellcolor[rgb]{ .71,  .827,  .851}0.05 & \cellcolor[rgb]{ .71,  .827,  .851}0.04 \bigstrut\\
\cline{3-10}          &       & FITNESS & \cellcolor[rgb]{ .435,  .627,  .675}0.77 & \cellcolor[rgb]{ .71,  .827,  .851}0.67 & \cellcolor[rgb]{ .435,  .627,  .675}0.74 & \cellcolor[rgb]{ .71,  .827,  .851}0.68 & \cellcolor[rgb]{ .435,  .627,  .675}0.03 & \cellcolor[rgb]{ .435,  .627,  .675}0.02 & \cellcolor[rgb]{ .435,  .627,  .675}0.01 \bigstrut\\
    \hline
    \end{tabular}%
    }
  \label{tab: exp_RQ1_1}%
  \vspace{-2ex} 
\end{table}%

%% file: exp_RQ1_2.tex
\begin{table}[htbp]
  \centering
  \caption{(RQ1) Results of comparing FITNESS with existing in/post-processing methods in different tasks.}
  \resizebox{\columnwidth}{!}{
    \begin{tabular}{|c|c|c|c|c|c|c|c|c|c|}
    \hline
    \rowcolor[rgb]{ .988,  .933,  .886} Alg.  & Task  & Method & Acc   & R     & P     & F1    & SPD   & AOD   & EOD \bigstrut\\
    \hline
    \multirow{15}[30]{*}{\begin{turn}{-90}Logistic Regression\end{turn}} & \multirow{5}[10]{*}{\begin{turn}{-90}Compas-Sex\end{turn}} & Default & \cellcolor[rgb]{ .435,  .627,  .675}0.67 & \cellcolor[rgb]{ .435,  .627,  .675}0.66 & \cellcolor[rgb]{ .435,  .627,  .675}0.67 & \cellcolor[rgb]{ .435,  .627,  .675}0.66 & 0.27  & 0.25  & 0.20 \bigstrut\\
\cline{3-10}          &       & ADV   & \cellcolor[rgb]{ .71,  .827,  .851}0.66 & \cellcolor[rgb]{ .71,  .827,  .851}0.65 & \cellcolor[rgb]{ .71,  .827,  .851}0.66 & \cellcolor[rgb]{ .71,  .827,  .851}0.65 & 0.28  & 0.26  & 0.20 \bigstrut\\
\cline{3-10}          &       & ROC   & \cellcolor[rgb]{ .71,  .827,  .851}0.66 & \cellcolor[rgb]{ .435,  .627,  .675}0.66 & \cellcolor[rgb]{ .71,  .827,  .851}0.66 & \cellcolor[rgb]{ .435,  .627,  .675}0.66 & \cellcolor[rgb]{ .435,  .627,  .675}0.04 & \cellcolor[rgb]{ .435,  .627,  .675}0.05 & \cellcolor[rgb]{ .435,  .627,  .675}0.04 \bigstrut\\
\cline{3-10}          &       & MAAT  & \cellcolor[rgb]{ .435,  .627,  .675}0.67 & \cellcolor[rgb]{ .71,  .827,  .851}0.65 & \cellcolor[rgb]{ .435,  .627,  .675}0.67 & \cellcolor[rgb]{ .71,  .827,  .851}0.65 & 0.17  & 0.14  & 0.10 \bigstrut\\
\cline{3-10}          &       & FITNESS & \cellcolor[rgb]{ .435,  .627,  .675}0.67 & \cellcolor[rgb]{ .435,  .627,  .675}0.66 & \cellcolor[rgb]{ .71,  .827,  .851}0.66 & \cellcolor[rgb]{ .435,  .627,  .675}0.66 & \cellcolor[rgb]{ .71,  .827,  .851}0.07 & \cellcolor[rgb]{ .435,  .627,  .675}0.05 & \cellcolor[rgb]{ .435,  .627,  .675}0.04 \bigstrut\\
\cline{2-10}          & \multirow{5}[10]{*}{\begin{turn}{-90}Compas-Race\end{turn}} & Default & \cellcolor[rgb]{ .435,  .627,  .675}0.67 & \cellcolor[rgb]{ .435,  .627,  .675}0.66 & \cellcolor[rgb]{ .435,  .627,  .675}0.67 & \cellcolor[rgb]{ .435,  .627,  .675}0.66 & 0.18  & 0.16  & 0.11 \bigstrut\\
\cline{3-10}          &       & ADV   & \cellcolor[rgb]{ .71,  .827,  .851}0.66 & \cellcolor[rgb]{ .71,  .827,  .851}0.65 & \cellcolor[rgb]{ .71,  .827,  .851}0.66 & \cellcolor[rgb]{ .71,  .827,  .851}0.65 & 0.13  & 0.11  & 0.09 \bigstrut\\
\cline{3-10}          &       & ROC   & \cellcolor[rgb]{ .71,  .827,  .851}0.66 & \cellcolor[rgb]{ .435,  .627,  .675}0.66 & \cellcolor[rgb]{ .71,  .827,  .851}0.66 & \cellcolor[rgb]{ .435,  .627,  .675}0.66 & \cellcolor[rgb]{ .435,  .627,  .675}0.03 & \cellcolor[rgb]{ .435,  .627,  .675}0.04 & \cellcolor[rgb]{ .71,  .827,  .851}0.04 \bigstrut\\
\cline{3-10}          &       & MAAT  & \cellcolor[rgb]{ .435,  .627,  .675}0.67 & \cellcolor[rgb]{ .435,  .627,  .675}0.66 & \cellcolor[rgb]{ .435,  .627,  .675}0.67 & \cellcolor[rgb]{ .435,  .627,  .675}0.66 & 0.09  & 0.07  & \cellcolor[rgb]{ .71,  .827,  .851}0.03 \bigstrut\\
\cline{3-10}          &       & FITNESS & \cellcolor[rgb]{ .435,  .627,  .675}0.67 & \cellcolor[rgb]{ .435,  .627,  .675}0.66 & \cellcolor[rgb]{ .435,  .627,  .675}0.67 & \cellcolor[rgb]{ .435,  .627,  .675}0.66 & \cellcolor[rgb]{ .71,  .827,  .851}0.05 & \cellcolor[rgb]{ .435,  .627,  .675}0.04 & \cellcolor[rgb]{ .435,  .627,  .675}0.02 \bigstrut\\
\cline{2-10}          & \multirow{5}[10]{*}{\begin{turn}{-90}Bank-Age\end{turn}} & Default & \cellcolor[rgb]{ .435,  .627,  .675}0.90 & \cellcolor[rgb]{ .71,  .827,  .851}0.68 & 0.79  & \cellcolor[rgb]{ .71,  .827,  .851}0.72 & 0.09  & 0.08  & 0.13 \bigstrut\\
\cline{3-10}          &       & ADV   & \cellcolor[rgb]{ .435,  .627,  .675}0.90 & \cellcolor[rgb]{ .435,  .627,  .675}0.72 & 0.78  & \cellcolor[rgb]{ .435,  .627,  .675}0.74 & 0.11  & 0.07  & 0.13 \bigstrut\\
\cline{3-10}          &       & ROC   & \cellcolor[rgb]{ .71,  .827,  .851}0.89 & 0.60  & \cellcolor[rgb]{ .435,  .627,  .675}0.81 & 0.62  & \cellcolor[rgb]{ .71,  .827,  .851}0.05 & 0.06  & 0.10 \bigstrut\\
\cline{3-10}          &       & MAAT  & \cellcolor[rgb]{ .435,  .627,  .675}0.90 & \cellcolor[rgb]{ .71,  .827,  .851}0.69 & 0.79  & \cellcolor[rgb]{ .71,  .827,  .851}0.72 & \cellcolor[rgb]{ .71,  .827,  .851}0.05 & \cellcolor[rgb]{ .435,  .627,  .675}0.03 & \cellcolor[rgb]{ .435,  .627,  .675}0.04 \bigstrut\\
\cline{3-10}          &       & FITNESS & \cellcolor[rgb]{ .435,  .627,  .675}0.90 & \cellcolor[rgb]{ .71,  .827,  .851}0.68 & \cellcolor[rgb]{ .71,  .827,  .851}0.80 & \cellcolor[rgb]{ .71,  .827,  .851}0.72 & \cellcolor[rgb]{ .435,  .627,  .675}0.04 & \cellcolor[rgb]{ .71,  .827,  .851}0.04 & \cellcolor[rgb]{ .71,  .827,  .851}0.06 \bigstrut\\
    \hline
    \end{tabular}%
    }
  \label{tab: exp_RQ1_2}%
  \vspace{-2ex} 
\end{table}%

%% file: exp_RQ1_3.tex
\begin{table}[htbp]
  \centering
  \caption{(RQ1) Results of comparing FITNESS with existing pre-processing methods with different ML algorithms.}
  \resizebox{\columnwidth}{!}{
    \begin{tabular}{|c|c|c|c|c|c|c|c|c|c|}
     \hline
    \rowcolor[rgb]{ .988,  .933,  .886} Task  & Alg.  & Method & Acc   & R     & P     & F1    & SPD   & AOD   & EOD \bigstrut\\
    \hline
    \multirow{15}[30]{*}{\begin{turn}{-90}Compas-Sex\end{turn}} & \multirow{5}[10]{*}{LR} & Default & \cellcolor[rgb]{ .435,  .627,  .675}0.67 & \cellcolor[rgb]{ .435,  .627,  .675}0.66 & \cellcolor[rgb]{ .435,  .627,  .675}0.67 & \cellcolor[rgb]{ .435,  .627,  .675}0.66 & 0.27  & 0.25  & 0.20 \bigstrut\\
\cline{3-10}          &       & REW   & \cellcolor[rgb]{ .71,  .827,  .851}0.66 & \cellcolor[rgb]{ .435,  .627,  .675}0.66 & \cellcolor[rgb]{ .71,  .827,  .851}0.66 & \cellcolor[rgb]{ .71,  .827,  .851}0.66 & \cellcolor[rgb]{ .435,  .627,  .675}0.05 & \cellcolor[rgb]{ .435,  .627,  .675}0.04 & \cellcolor[rgb]{ .435,  .627,  .675}0.03 \bigstrut\\
\cline{3-10}          &       & Fairway & \cellcolor[rgb]{ .71,  .827,  .851}0.66 & \cellcolor[rgb]{ .71,  .827,  .851}0.65 & \cellcolor[rgb]{ .71,  .827,  .851}0.66 & \cellcolor[rgb]{ .71,  .827,  .851}0.64 & 0.20  & 0.17  & 0.12 \bigstrut\\
\cline{3-10}          &       & Fair SMOTE & \cellcolor[rgb]{ .71,  .827,  .851}0.65 & \cellcolor[rgb]{ .71,  .827,  .851}0.65 & \cellcolor[rgb]{ .71,  .827,  .851}0.65 & \cellcolor[rgb]{ .71,  .827,  .851}0.65 & 0.09  & 0.07  & \cellcolor[rgb]{ .435,  .627,  .675}0.04 \bigstrut\\
\cline{3-10}          &       & FITNESS & \cellcolor[rgb]{ .435,  .627,  .675}0.67 & \cellcolor[rgb]{ .435,  .627,  .675}0.66 & \cellcolor[rgb]{ .71,  .827,  .851}0.66 & \cellcolor[rgb]{ .435,  .627,  .675}0.66 & \cellcolor[rgb]{ .71,  .827,  .851}0.07 & \cellcolor[rgb]{ .435,  .627,  .675}0.05 & \cellcolor[rgb]{ .435,  .627,  .675}0.04 \bigstrut\\
\cline{2-10}          & \multirow{5}[10]{*}{RF} & Default & \cellcolor[rgb]{ .71,  .827,  .851}0.65 & \cellcolor[rgb]{ .71,  .827,  .851}0.64 & \cellcolor[rgb]{ .71,  .827,  .851}0.64 & \cellcolor[rgb]{ .71,  .827,  .851}0.64 & 0.17  & 0.14  & 0.12 \bigstrut\\
\cline{3-10}          &       & REW   & \cellcolor[rgb]{ .71,  .827,  .851}0.65 & \cellcolor[rgb]{ .71,  .827,  .851}0.64 & \cellcolor[rgb]{ .71,  .827,  .851}0.65 & \cellcolor[rgb]{ .71,  .827,  .851}0.65 & 0.15  & 0.12  & 0.11 \bigstrut\\
\cline{3-10}          &       & Fairway & \cellcolor[rgb]{ .435,  .627,  .675}0.67 & \cellcolor[rgb]{ .435,  .627,  .675}0.66 & \cellcolor[rgb]{ .435,  .627,  .675}0.68 & \cellcolor[rgb]{ .435,  .627,  .675}0.66 & 0.14  & 0.11  & 0.08 \bigstrut\\
\cline{3-10}          &       & Fair SMOTE & \cellcolor[rgb]{ .435,  .627,  .675}0.67 & \cellcolor[rgb]{ .435,  .627,  .675}0.66 & \cellcolor[rgb]{ .71,  .827,  .851}0.66 & \cellcolor[rgb]{ .435,  .627,  .675}0.66 & 0.11  & 0.08  & 0.06 \bigstrut\\
\cline{3-10}          &       & FITNESS & \cellcolor[rgb]{ .71,  .827,  .851}0.65 & \cellcolor[rgb]{ .71,  .827,  .851}0.64 & \cellcolor[rgb]{ .71,  .827,  .851}0.64 & \cellcolor[rgb]{ .71,  .827,  .851}0.64 & \cellcolor[rgb]{ .435,  .627,  .675}0.05 & \cellcolor[rgb]{ .435,  .627,  .675}0.03 & \cellcolor[rgb]{ .435,  .627,  .675}0.02 \bigstrut\\
\cline{2-10}          & \multirow{5}[10]{*}{SVM} & Default & \cellcolor[rgb]{ .435,  .627,  .675}0.66 & \cellcolor[rgb]{ .435,  .627,  .675}0.66 & \cellcolor[rgb]{ .435,  .627,  .675}0.66 & \cellcolor[rgb]{ .435,  .627,  .675}0.66 & 0.26  & 0.24  & 0.18 \bigstrut\\
\cline{3-10}          &       & REW   & \cellcolor[rgb]{ .435,  .627,  .675}0.66 & \cellcolor[rgb]{ .71,  .827,  .851}0.65 & \cellcolor[rgb]{ .435,  .627,  .675}0.66 & \cellcolor[rgb]{ .71,  .827,  .851}0.65 & \cellcolor[rgb]{ .435,  .627,  .675}0.06 & \cellcolor[rgb]{ .435,  .627,  .675}0.05 & \cellcolor[rgb]{ .435,  .627,  .675}0.03 \bigstrut\\
\cline{3-10}          &       & Fairway & 0.62  & 0.61  & \cellcolor[rgb]{ .71,  .827,  .851}0.64 & 0.60  & 0.16  & 0.15  & 0.12 \bigstrut\\
\cline{3-10}          &       & Fair SMOTE & \cellcolor[rgb]{ .71,  .827,  .851}0.65 & \cellcolor[rgb]{ .71,  .827,  .851}0.64 & \cellcolor[rgb]{ .71,  .827,  .851}0.64 & \cellcolor[rgb]{ .71,  .827,  .851}0.64 & 0.09  & 0.06  & \cellcolor[rgb]{ .71,  .827,  .851}0.04 \bigstrut\\
\cline{3-10}          &       & FITNESS & \cellcolor[rgb]{ .435,  .627,  .675}0.66 & \cellcolor[rgb]{ .71,  .827,  .851}0.65 & \cellcolor[rgb]{ .435,  .627,  .675}0.66 & \cellcolor[rgb]{ .71,  .827,  .851}0.65 & \cellcolor[rgb]{ .435,  .627,  .675}0.06 & \cellcolor[rgb]{ .435,  .627,  .675}0.05 & \cellcolor[rgb]{ .435,  .627,  .675}0.03 \bigstrut\\
    \hline
    
    \end{tabular}%
    }
  \label{tab: exp_RQ1_3}%
  \vspace{-2ex} 
\end{table}%

%% file: exp_RQ1_4.tex
\begin{table}[htbp]
  \centering
  \caption{(RQ1) Results of comparing FITNESS with existing in/post-processing methods with different ML algorithms.}
  \resizebox{\columnwidth}{!}{
    \begin{tabular}{|c|c|c|c|c|c|c|c|c|c|}

    \hline
    \rowcolor[rgb]{ .988,  .933,  .886} Task  & Alg.  & Method & Acc   & R     & P     & F1    & SPD   & AOD   & EOD \bigstrut\\
    \hline
    \multirow{15}[30]{*}{\begin{turn}{-90}Adult-Sex\end{turn}} & \multirow{5}[10]{*}{LR} & Default & \cellcolor[rgb]{ .435,  .627,  .675}0.85 & \cellcolor[rgb]{ .71,  .827,  .851}0.76 & \cellcolor[rgb]{ .71,  .827,  .851}0.80 & \cellcolor[rgb]{ .435,  .627,  .675}0.78 & 0.19  & 0.10  & 0.12 \bigstrut\\
\cline{3-10}          &       & ADV   & \cellcolor[rgb]{ .71,  .827,  .851}0.83 & 0.72  & \cellcolor[rgb]{ .71,  .827,  .851}0.80 & 0.75  & \cellcolor[rgb]{ .435,  .627,  .675}0.03 & 0.14  & 0.27 \bigstrut\\
\cline{3-10}          &       & ROC   & 0.79  & \cellcolor[rgb]{ .435,  .627,  .675}0.79 & 0.73  & 0.75  & \cellcolor[rgb]{ .71,  .827,  .851}0.05 & 0.11  & 0.16 \bigstrut\\
\cline{3-10}          &       & MAAT  & \cellcolor[rgb]{ .71,  .827,  .851}0.84 & 0.72  & \cellcolor[rgb]{ .435,  .627,  .675}0.82 & 0.75  & 0.11  & \cellcolor[rgb]{ .71,  .827,  .851}0.04 & \cellcolor[rgb]{ .71,  .827,  .851}0.04 \bigstrut\\
\cline{3-10}          &       & FITNESS & \cellcolor[rgb]{ .71,  .827,  .851}0.84 & \cellcolor[rgb]{ .71,  .827,  .851}0.75 & \cellcolor[rgb]{ .71,  .827,  .851}0.80 & \cellcolor[rgb]{ .71,  .827,  .851}0.77 & 0.14  & \cellcolor[rgb]{ .435,  .627,  .675}0.03 & \cellcolor[rgb]{ .435,  .627,  .675}0.01 \bigstrut\\
\cline{2-10}          & \multirow{5}[10]{*}{RF} & Default & \cellcolor[rgb]{ .435,  .627,  .675}0.84 & \cellcolor[rgb]{ .435,  .627,  .675}0.77 & \cellcolor[rgb]{ .71,  .827,  .851}0.79 & \cellcolor[rgb]{ .435,  .627,  .675}0.78 & 0.19  & 0.08  & 0.08 \bigstrut\\
\cline{3-10}          &       & ADV   & \cellcolor[rgb]{ .71,  .827,  .851}0.83 & \cellcolor[rgb]{ .71,  .827,  .851}0.72 & \cellcolor[rgb]{ .71,  .827,  .851}0.80 & \cellcolor[rgb]{ .71,  .827,  .851}0.75 & \cellcolor[rgb]{ .435,  .627,  .675}0.03 & 0.14  & 0.27 \bigstrut\\
\cline{3-10}          &       & ROC   & 0.77  & \cellcolor[rgb]{ .71,  .827,  .851}0.75 & 0.71  & \cellcolor[rgb]{ .71,  .827,  .851}0.73 & \cellcolor[rgb]{ .71,  .827,  .851}0.06 & 0.20  & 0.26 \bigstrut\\
\cline{3-10}          &       & MAAT  & \cellcolor[rgb]{ .435,  .627,  .675}0.84 & \cellcolor[rgb]{ .71,  .827,  .851}0.74 & \cellcolor[rgb]{ .435,  .627,  .675}0.81 & \cellcolor[rgb]{ .71,  .827,  .851}0.77 & 0.12  & \cellcolor[rgb]{ .71,  .827,  .851}0.04 & \cellcolor[rgb]{ .71,  .827,  .851}0.04 \bigstrut\\
\cline{3-10}          &       & FITNESS & \cellcolor[rgb]{ .435,  .627,  .675}0.84 & \cellcolor[rgb]{ .71,  .827,  .851}0.73 & \cellcolor[rgb]{ .435,  .627,  .675}0.81 & \cellcolor[rgb]{ .71,  .827,  .851}0.76 & 0.12  & \cellcolor[rgb]{ .435,  .627,  .675}0.02 & \cellcolor[rgb]{ .435,  .627,  .675}0.00 \bigstrut\\
\cline{2-10}          & \multirow{5}[10]{*}{SVM} & Default & \cellcolor[rgb]{ .435,  .627,  .675}0.85 & \cellcolor[rgb]{ .71,  .827,  .851}0.76 & \cellcolor[rgb]{ .71,  .827,  .851}0.81 & \cellcolor[rgb]{ .435,  .627,  .675}0.78 & 0.18  & 0.08  & 0.09 \bigstrut\\
\cline{3-10}          &       & ADV   & \cellcolor[rgb]{ .71,  .827,  .851}0.83 & 0.72  & \cellcolor[rgb]{ .71,  .827,  .851}0.80 & \cellcolor[rgb]{ .71,  .827,  .851}0.75 & \cellcolor[rgb]{ .435,  .627,  .675}0.03 & 0.14  & 0.27 \bigstrut\\
\cline{3-10}          &       & ROC   & 0.78  & \cellcolor[rgb]{ .435,  .627,  .675}0.79 & 0.73  & \cellcolor[rgb]{ .71,  .827,  .851}0.75 & \cellcolor[rgb]{ .71,  .827,  .851}0.05 & 0.11  & 0.16 \bigstrut\\
\cline{3-10}          &       & MAAT  & \cellcolor[rgb]{ .71,  .827,  .851}0.84 & 0.72  & \cellcolor[rgb]{ .435,  .627,  .675}0.82 & \cellcolor[rgb]{ .71,  .827,  .851}0.75 & 0.09  & \cellcolor[rgb]{ .71,  .827,  .851}0.05 & 0.09 \bigstrut\\
\cline{3-10}          &       & FITNESS & \cellcolor[rgb]{ .435,  .627,  .675}0.85 & \cellcolor[rgb]{ .71,  .827,  .851}0.76 & \cellcolor[rgb]{ .71,  .827,  .851}0.81 & \cellcolor[rgb]{ .435,  .627,  .675}0.78 & 0.15  & \cellcolor[rgb]{ .435,  .627,  .675}0.03 & \cellcolor[rgb]{ .435,  .627,  .675}0.01 \bigstrut\\
    \hline
    \end{tabular}%
    }
  \label{tab: exp_RQ1_4}%
  \vspace{-2ex} 
\end{table}%

%% file: exp_RQ3.tex
\begin{table}[htbp]
  \centering
  \caption{(RQ3) Results of comparing FITNESS with existing methods in protecting multiple sensitive features.}
  \resizebox{\columnwidth}{!}{
    \begin{tabular}{|c|c|c|c|c|c|c|c|c|c|}
    \hline
    \rowcolor[rgb]{ .988,  .933,  .886} Alg.  & Task  & Method & Acc   & R     & P     & F1    & SPD   & AOD   & EOD \bigstrut\\
    \hline
    \multirow{20}[40]{*}{\begin{turn}{-90}Random Forest\end{turn}} & \multirow{5}[10]{*}{\begin{turn}{-90}Adult-Sex\end{turn}} & Default & \cellcolor[rgb]{ .71,  .827,  .851}0.84 & \cellcolor[rgb]{ .71,  .827,  .851}0.77 & \cellcolor[rgb]{ .71,  .827,  .851}0.80 & \cellcolor[rgb]{ .71,  .827,  .851}0.78 & 0.19  & 0.08  & 0.08 \bigstrut\\
\cline{3-10}          &       & MAAT  & \cellcolor[rgb]{ .435,  .627,  .675}0.85 & \cellcolor[rgb]{ .71,  .827,  .851}0.74 & \cellcolor[rgb]{ .71,  .827,  .851}0.81 & \cellcolor[rgb]{ .71,  .827,  .851}0.77 & 0.13  & \cellcolor[rgb]{ .71,  .827,  .851}0.03 & \cellcolor[rgb]{ .435,  .627,  .675}0.02 \bigstrut\\
\cline{3-10}          &       & Fair SMOTE & \cellcolor[rgb]{ .71,  .827,  .851}0.83 & \cellcolor[rgb]{ .435,  .627,  .675}0.78 & \cellcolor[rgb]{ .71,  .827,  .851}0.78 & \cellcolor[rgb]{ .71,  .827,  .851}0.78 & 0.22  & 0.11  & 0.10 \bigstrut\\
\cline{3-10}          &       & xFAIR & \cellcolor[rgb]{ .71,  .827,  .851}0.83 & 0.52  & 0.69  & 0.59  & \cellcolor[rgb]{ .435,  .627,  .675}0.11 & \cellcolor[rgb]{ .435,  .627,  .675}0.02 & 0.06 \bigstrut\\
\cline{3-10}          &       & FITNESS & \cellcolor[rgb]{ .71,  .827,  .851}0.84 & \cellcolor[rgb]{ .71,  .827,  .851}0.73 & \cellcolor[rgb]{ .71,  .827,  .851}0.81 & \cellcolor[rgb]{ .71,  .827,  .851}0.76 & \cellcolor[rgb]{ .71,  .827,  .851}0.12 & \cellcolor[rgb]{ .71,  .827,  .851}0.03 & \cellcolor[rgb]{ .71,  .827,  .851}0.03 \bigstrut\\
\cline{2-10}          & \multirow{5}[10]{*}{\begin{turn}{-90}Adult-Race\end{turn}} & Default & \cellcolor[rgb]{ .71,  .827,  .851}0.84 & \cellcolor[rgb]{ .71,  .827,  .851}0.77 & \cellcolor[rgb]{ .71,  .827,  .851}0.80 & \cellcolor[rgb]{ .71,  .827,  .851}0.78 & 0.10  & 0.05  & 0.04 \bigstrut\\
\cline{3-10}          &       & MAAT  & \cellcolor[rgb]{ .435,  .627,  .675}0.85 & \cellcolor[rgb]{ .71,  .827,  .851}0.74 & \cellcolor[rgb]{ .71,  .827,  .851}0.81 & \cellcolor[rgb]{ .71,  .827,  .851}0.77 & 0.07  & \cellcolor[rgb]{ .71,  .827,  .851}0.02 & \cellcolor[rgb]{ .71,  .827,  .851}0.03 \bigstrut\\
\cline{3-10}          &       & Fair SMOTE & \cellcolor[rgb]{ .71,  .827,  .851}0.83 & \cellcolor[rgb]{ .435,  .627,  .675}0.78 & \cellcolor[rgb]{ .71,  .827,  .851}0.78 & \cellcolor[rgb]{ .71,  .827,  .851}0.78 & 0.13  & 0.07  & 0.08 \bigstrut\\
\cline{3-10}          &       & xFAIR & \cellcolor[rgb]{ .71,  .827,  .851}0.83 & 0.52  & 0.69  & 0.59  & \cellcolor[rgb]{ .435,  .627,  .675}0.06 & \cellcolor[rgb]{ .435,  .627,  .675}0.00 & \cellcolor[rgb]{ .435,  .627,  .675}0.02 \bigstrut\\
\cline{3-10}          &       & FITNESS & \cellcolor[rgb]{ .71,  .827,  .851}0.84 & \cellcolor[rgb]{ .71,  .827,  .851}0.73 & \cellcolor[rgb]{ .71,  .827,  .851}0.81 & \cellcolor[rgb]{ .71,  .827,  .851}0.76 & \cellcolor[rgb]{ .435,  .627,  .675}0.06 & \cellcolor[rgb]{ .71,  .827,  .851}0.02 & \cellcolor[rgb]{ .71,  .827,  .851}0.03 \bigstrut\\
\cline{2-10}          & \multirow{5}[10]{*}{\begin{turn}{-90}Compas-Sex\end{turn}} & Default & \cellcolor[rgb]{ .435,  .627,  .675}0.65 & \cellcolor[rgb]{ .71,  .827,  .851}0.64 & \cellcolor[rgb]{ .71,  .827,  .851}0.64 & \cellcolor[rgb]{ .71,  .827,  .851}0.64 & 0.17  & 0.14  & 0.12 \bigstrut\\
\cline{3-10}          &       & MAAT  & \cellcolor[rgb]{ .435,  .627,  .675}0.65 & \cellcolor[rgb]{ .71,  .827,  .851}0.64 & \cellcolor[rgb]{ .71,  .827,  .851}0.65 & \cellcolor[rgb]{ .71,  .827,  .851}0.64 & 0.13  & 0.10  & 0.08 \bigstrut\\
\cline{3-10}          &       & Fair SMOTE & \cellcolor[rgb]{ .71,  .827,  .851}0.64 & \cellcolor[rgb]{ .71,  .827,  .851}0.64 & \cellcolor[rgb]{ .71,  .827,  .851}0.64 & \cellcolor[rgb]{ .71,  .827,  .851}0.64 & 0.17  & 0.14  & 0.11 \bigstrut\\
\cline{3-10}          &       & xFAIR & \cellcolor[rgb]{ .435,  .627,  .675}0.65 & \cellcolor[rgb]{ .435,  .627,  .675}0.73 & \cellcolor[rgb]{ .435,  .627,  .675}0.67 & \cellcolor[rgb]{ .435,  .627,  .675}0.70 & \cellcolor[rgb]{ .71,  .827,  .851}0.09 & \cellcolor[rgb]{ .435,  .627,  .675}0.01 & \cellcolor[rgb]{ .71,  .827,  .851}0.05 \bigstrut\\
\cline{3-10}          &       & FITNESS & \cellcolor[rgb]{ .435,  .627,  .675}0.65 & \cellcolor[rgb]{ .71,  .827,  .851}0.64 & \cellcolor[rgb]{ .71,  .827,  .851}0.64 & \cellcolor[rgb]{ .71,  .827,  .851}0.64 & \cellcolor[rgb]{ .435,  .627,  .675}0.07 & \cellcolor[rgb]{ .71,  .827,  .851}0.04 & \cellcolor[rgb]{ .71,  .827,  .851}0.04 \bigstrut\\
\cline{2-10}          & \multirow{5}[10]{*}{\begin{turn}{-90}Compas-Race\end{turn}} & Default & \cellcolor[rgb]{ .435,  .627,  .675}0.65 & \cellcolor[rgb]{ .71,  .827,  .851}0.64 & \cellcolor[rgb]{ .71,  .827,  .851}0.64 & \cellcolor[rgb]{ .71,  .827,  .851}0.64 & 0.14  & 0.12  & 0.09 \bigstrut\\
\cline{3-10}          &       & MAAT  & \cellcolor[rgb]{ .435,  .627,  .675}0.65 & \cellcolor[rgb]{ .71,  .827,  .851}0.64 & \cellcolor[rgb]{ .71,  .827,  .851}0.65 & \cellcolor[rgb]{ .71,  .827,  .851}0.64 & 0.11  & 0.09  & \cellcolor[rgb]{ .71,  .827,  .851}0.06 \bigstrut\\
\cline{3-10}          &       & Fair SMOTE & \cellcolor[rgb]{ .71,  .827,  .851}0.64 & \cellcolor[rgb]{ .71,  .827,  .851}0.64 & \cellcolor[rgb]{ .71,  .827,  .851}0.64 & \cellcolor[rgb]{ .71,  .827,  .851}0.64 & 0.12  & 0.10  & 0.07 \bigstrut\\
\cline{3-10}          &       & xFAIR & \cellcolor[rgb]{ .435,  .627,  .675}0.65 & \cellcolor[rgb]{ .435,  .627,  .675}0.73 & \cellcolor[rgb]{ .435,  .627,  .675}0.67 & \cellcolor[rgb]{ .435,  .627,  .675}0.70 & 0.13  & \cellcolor[rgb]{ .435,  .627,  .675}0.02 & 0.09 \bigstrut\\
\cline{3-10}          &       & FITNESS & \cellcolor[rgb]{ .435,  .627,  .675}0.65 & \cellcolor[rgb]{ .71,  .827,  .851}0.64 & \cellcolor[rgb]{ .71,  .827,  .851}0.64 & \cellcolor[rgb]{ .71,  .827,  .851}0.64 & \cellcolor[rgb]{ .435,  .627,  .675}0.04 & \cellcolor[rgb]{ .71,  .827,  .851}0.04 & \cellcolor[rgb]{ .435,  .627,  .675}0.03 \bigstrut\\
    \hline
    \end{tabular}%
    }
  \label{tab: exp_RQ3}%
  \vspace{-2ex} 
\end{table}%

%% file: 6.discussion.tex
\section{DISCUSSION}
\label{sec:dis}
Here we discuss the strength, threat of effectiveness of FITNESS.
\subsection{Why FITNESS?}
\textbf{Low Performance Impact.} 
FITNESS does not enhance model fairness by sacrificing model performance as compared with previous work \cite{kamiran2012data, kamiran2012decision, chakraborty2020fairway, chakraborty2021bias, chen2022maat, zhang2018mitigating, peng2021xfair}, FITNESS achieves the best fairness improvement and minimum performance decreases among all our experiments.

\noindent\textbf{Applicability.} 
As a pre-processing method that reduces training data bias, FITNESS does not require mitigating bias in every time the model is used, unlike in-processing and post-processing methods. It also does not impact the use of the original data in downstream applications. Furthermore, the use of different ML algorithms does not affect the validity of FITNESS. This makes it a practical and widely applicable solution for mitigating bias.

\noindent\textbf{Flexibility.} 
Because of multi-objective optimization, FITNESS is more flexible than other methods. This optimization technique allows developers to adjust the composition and weights of the optimization function freely, enabling targeted optimization of the machine learning software. This added flexibility makes FITNESS applicable in a broader range of scenarios.


\subsection{Threats to Validity}

\noindent \textbf{Evaluation Bias.}
 The selection of benchmark datasets, tasks, evaluation criteria, and existing methods may threaten the validity of our results. To mitigate these threats, we follow previous work \cite{chakraborty2020fairway, chakraborty2021bias, chen2022maat, peng2021xfair} and adopt widely used benchmark datasets, and multiple metrics. Still, the results may change a bit if new datasets, algorithms and evaluation metrics are utilized.

 \noindent \textbf{Selection of ML algorithms.}
As a pre-processing approach mitigating bias by modifying training data, FITNESS can apply to any ML algorithm, including deep learning (DL) algorithms. However, as DL algorithms are good at processing complex unstructured  (\eg images) and easily overfit in small tabular data tasks \cite{chen2022maat}, following state-of-the-art studies \cite{chakraborty2020fairway, chakraborty2021bias, chen2022maat}, we adopt 3 broadly used classic ML algorithms in our research. One could replicate our work with DL algorithms in the future.

\noindent \textbf{Internal Validity.}
The existing bias-mitigating approaches, including FITNESS, require full access to the training data to process the sensitive features. If policies (\eg General Data Protection Regulation, GDPR \cite{voigt2017eu}) of protecting personal privacy prevent developer access to such information, the validity of FITNESS will be affected. Still, FITNESS can be applied to most scenarios. In the future, we would like to explore the bias-mitigating methods that need not access such sensitive information.

\noindent \textbf{External Validity.}
This work focuses on classification tasks and tabular data, and it cannot mitigate bias in data mining and image processing as they are out of the scope of this study.

%% file: 7.conclusion.tex
\section{CONCLUSION}
\label{sec:conc}
This paper proposed FITNESS, a data and model biases mitigation approach that can significantly boost ML software fairness without decreasing too much prediction performance. To our knowledge, FITNESS is the first method that alleviates bias via causality analysis and de-correlation. As a pre-processing approach, FITNESS can work independently for training discrimination-free models or be combined with other strategies to obtain unbiased datasets in various scenarios. In a nutshell, we devise an effective approach to address a pain point issue in machine learning software. Furthermore, the successful practice of de-correlation and multi-objective optimization in FITNESS opened up potential research opportunities in the AI and SE communities, including balancing ML software robustness, efficiency, privacy and performance.

%% file: 8.availability.tex
\section{Data Availability}
\label{sec:relatedWork}
To facilitate future work and replication of our approach, we make the replication package of FITNESS, including code, datasets, and other supplementary materials available on the website \cite{doiReplicationPackage}.

%% file: main.bbl

\begin{thebibliography}{70}


\ifx \showCODEN    \undefined \def \showCODEN     #1{\unskip}     \fi
\ifx \showDOI      \undefined \def \showDOI       #1{#1}\fi
\ifx \showISBNx    \undefined \def \showISBNx     #1{\unskip}     \fi
\ifx \showISBNxiii \undefined \def \showISBNxiii  #1{\unskip}     \fi
\ifx \showISSN     \undefined \def \showISSN      #1{\unskip}     \fi
\ifx \showLCCN     \undefined \def \showLCCN      #1{\unskip}     \fi
\ifx \shownote     \undefined \def \shownote      #1{#1}          \fi
\ifx \showarticletitle \undefined \def \showarticletitle #1{#1}   \fi
\ifx \showURL      \undefined \def \showURL       {\relax}        \fi
\providecommand\bibfield[2]{#2}
\providecommand\bibinfo[2]{#2}
\providecommand\natexlab[1]{#1}
\providecommand\showeprint[2][]{arXiv:#2}

\bibitem[\protect\citeauthoryear{??}{mis}{1996}]%
        {misc_adult_2}
 \bibinfo{year}{1996}\natexlab{}.
\newblock \bibinfo{title}{{The Adult Census Income dataset}}.
\newblock \bibinfo{howpublished}{UCI Machine Learning Repository}.
\newblock
\newblock
\shownote{https://archive.ics.uci.edu/ml/datasets/adult.}


\bibitem[\protect\citeauthoryear{??}{mis}{2019}]%
        {misc_south_german_credit_522}
 \bibinfo{year}{2019}\natexlab{}.
\newblock \bibinfo{title}{{The South German Credit dataset}}.
\newblock \bibinfo{howpublished}{UCI Machine Learning Repository}.
\newblock
\newblock
\shownote{https://archive.ics.uci.edu/ml/datasets/statlog+(german+credit+data).}


\bibitem[\protect\citeauthoryear{??}{mis}{2021}]%
        {misc_compas_dataset}
 \bibinfo{year}{2021}\natexlab{}.
\newblock \bibinfo{title}{{The Compas dataset}}.
\newblock \bibinfo{howpublished}{ProPublica}.
\newblock
\newblock
\shownote{https://github.com/propublica/compas-analysis.}


\bibitem[\protect\citeauthoryear{Aggarwal, Lohia, Nagar, Dey, and
  Saha}{Aggarwal et~al\mbox{.}}{2019}]%
        {aggarwal2019black}
\bibfield{author}{\bibinfo{person}{Aniya Aggarwal}, \bibinfo{person}{Pranay
  Lohia}, \bibinfo{person}{Seema Nagar}, \bibinfo{person}{Kuntal Dey}, {and}
  \bibinfo{person}{Diptikalyan Saha}.} \bibinfo{year}{2019}\natexlab{}.
\newblock \showarticletitle{Black box fairness testing of machine learning
  models}. In \bibinfo{booktitle}{\emph{Proceedings of the 2019 27th ACM Joint
  Meeting on European Software Engineering Conference and Symposium on the
  Foundations of Software Engineering}}. \bibinfo{pages}{625--635}.
\newblock


\bibitem[\protect\citeauthoryear{Anonymous}{Anonymous}{[n.d.]}]%
        {doiReplicationPackage}
\bibfield{author}{\bibinfo{person}{Anonymous}.}
  \bibinfo{year}{[n.d.]}\natexlab{}.
\newblock \bibinfo{title}{{R}eplication {P}ackage, {E}xperiments {R}esults and
  {S}upplementary {M}aterial of {F}{I}{T}{N}{E}{S}{S} --- doi.org}.
\newblock \bibinfo{howpublished}{\url{https://doi.org/10.5281/zenodo.7608487}}.
\newblock
\newblock
\shownote{[Accessed 06-Feb-2023].}


\bibitem[\protect\citeauthoryear{Assembly et~al\mbox{.}}{Assembly
  et~al\mbox{.}}{1948}]%
        {assembly1948universal}
\bibfield{author}{\bibinfo{person}{UN~General Assembly} {et~al\mbox{.}}}
  \bibinfo{year}{1948}\natexlab{}.
\newblock \showarticletitle{Universal declaration of human rights}.
\newblock \bibinfo{journal}{\emph{UN General Assembly}} \bibinfo{volume}{302},
  \bibinfo{number}{2} (\bibinfo{year}{1948}), \bibinfo{pages}{14--25}.
\newblock


\bibitem[\protect\citeauthoryear{Bellamy, Dey, Hind, Hoffman, Houde, Kannan,
  Lohia, Martino, Mehta, Mojsilovi{\'c}, et~al\mbox{.}}{Bellamy
  et~al\mbox{.}}{2019}]%
        {bellamy2019ai}
\bibfield{author}{\bibinfo{person}{Rachel~KE Bellamy}, \bibinfo{person}{Kuntal
  Dey}, \bibinfo{person}{Michael Hind}, \bibinfo{person}{Samuel~C Hoffman},
  \bibinfo{person}{Stephanie Houde}, \bibinfo{person}{Kalapriya Kannan},
  \bibinfo{person}{Pranay Lohia}, \bibinfo{person}{Jacquelyn Martino},
  \bibinfo{person}{Sameep Mehta}, \bibinfo{person}{Aleksandra Mojsilovi{\'c}},
  {et~al\mbox{.}}} \bibinfo{year}{2019}\natexlab{}.
\newblock \showarticletitle{AI Fairness 360: An extensible toolkit for
  detecting and mitigating algorithmic bias}.
\newblock \bibinfo{journal}{\emph{IBM Journal of Research and Development}}
  \bibinfo{volume}{63}, \bibinfo{number}{4/5} (\bibinfo{year}{2019}),
  \bibinfo{pages}{4--1}.
\newblock


\bibitem[\protect\citeauthoryear{Beutel, Chen, Doshi, Qian, Wei, Wu, Heldt,
  Zhao, Hong, Chi, et~al\mbox{.}}{Beutel et~al\mbox{.}}{2019}]%
        {beutel2019fairness}
\bibfield{author}{\bibinfo{person}{Alex Beutel}, \bibinfo{person}{Jilin Chen},
  \bibinfo{person}{Tulsee Doshi}, \bibinfo{person}{Hai Qian},
  \bibinfo{person}{Li Wei}, \bibinfo{person}{Yi Wu}, \bibinfo{person}{Lukasz
  Heldt}, \bibinfo{person}{Zhe Zhao}, \bibinfo{person}{Lichan Hong},
  \bibinfo{person}{Ed~H Chi}, {et~al\mbox{.}}} \bibinfo{year}{2019}\natexlab{}.
\newblock \showarticletitle{Fairness in recommendation ranking through pairwise
  comparisons}. In \bibinfo{booktitle}{\emph{Proceedings of the 25th ACM SIGKDD
  International Conference on Knowledge Discovery \& Data Mining}}.
  \bibinfo{pages}{2212--2220}.
\newblock


\bibitem[\protect\citeauthoryear{Biau and Scornet}{Biau and Scornet}{2016}]%
        {biau2016random}
\bibfield{author}{\bibinfo{person}{G{\'e}rard Biau} {and}
  \bibinfo{person}{Erwan Scornet}.} \bibinfo{year}{2016}\natexlab{}.
\newblock \showarticletitle{A random forest guided tour}.
\newblock \bibinfo{journal}{\emph{Test}} \bibinfo{volume}{25},
  \bibinfo{number}{2} (\bibinfo{year}{2016}), \bibinfo{pages}{197--227}.
\newblock


\bibitem[\protect\citeauthoryear{Bird, Dud{\'\i}k, Edgar, Horn, Lutz, Milan,
  Sameki, Wallach, and Walker}{Bird et~al\mbox{.}}{2020}]%
        {bird2020fairlearn}
\bibfield{author}{\bibinfo{person}{Sarah Bird}, \bibinfo{person}{Miro
  Dud{\'\i}k}, \bibinfo{person}{Richard Edgar}, \bibinfo{person}{Brandon Horn},
  \bibinfo{person}{Roman Lutz}, \bibinfo{person}{Vanessa Milan},
  \bibinfo{person}{Mehrnoosh Sameki}, \bibinfo{person}{Hanna Wallach}, {and}
  \bibinfo{person}{Kathleen Walker}.} \bibinfo{year}{2020}\natexlab{}.
\newblock \showarticletitle{Fairlearn: A toolkit for assessing and improving
  fairness in AI}.
\newblock \bibinfo{journal}{\emph{Microsoft, Tech. Rep. MSR-TR-2020-32}}
  (\bibinfo{year}{2020}).
\newblock


\bibitem[\protect\citeauthoryear{Biswas and Rajan}{Biswas and Rajan}{2020}]%
        {biswas2020machine}
\bibfield{author}{\bibinfo{person}{Sumon Biswas} {and} \bibinfo{person}{Hridesh
  Rajan}.} \bibinfo{year}{2020}\natexlab{}.
\newblock \showarticletitle{Do the machine learning models on a crowd sourced
  platform exhibit bias? an empirical study on model fairness}. In
  \bibinfo{booktitle}{\emph{Proceedings of the 28th ACM joint meeting on
  European software engineering conference and symposium on the foundations of
  software engineering}}. \bibinfo{pages}{642--653}.
\newblock


\bibitem[\protect\citeauthoryear{Bunge}{Bunge}{2017}]%
        {bunge2017causality}
\bibfield{author}{\bibinfo{person}{Mario Bunge}.}
  \bibinfo{year}{2017}\natexlab{}.
\newblock \bibinfo{booktitle}{\emph{Causality and modern science}}.
\newblock \bibinfo{publisher}{Routledge}.
\newblock


\bibitem[\protect\citeauthoryear{Censor}{Censor}{1977}]%
        {censor1977pareto}
\bibfield{author}{\bibinfo{person}{Yair Censor}.}
  \bibinfo{year}{1977}\natexlab{}.
\newblock \showarticletitle{Pareto optimality in multiobjective problems}.
\newblock \bibinfo{journal}{\emph{Applied Mathematics and Optimization}}
  \bibinfo{volume}{4}, \bibinfo{number}{1} (\bibinfo{year}{1977}),
  \bibinfo{pages}{41--59}.
\newblock


\bibitem[\protect\citeauthoryear{Chakraborty, Majumder, and
  Menzies}{Chakraborty et~al\mbox{.}}{2021}]%
        {chakraborty2021bias}
\bibfield{author}{\bibinfo{person}{Joymallya Chakraborty},
  \bibinfo{person}{Suvodeep Majumder}, {and} \bibinfo{person}{Tim Menzies}.}
  \bibinfo{year}{2021}\natexlab{}.
\newblock \showarticletitle{Bias in machine learning software: why? how? what
  to do?}. In \bibinfo{booktitle}{\emph{Proceedings of the 29th ACM Joint
  Meeting on European Software Engineering Conference and Symposium on the
  Foundations of Software Engineering}}. \bibinfo{pages}{429--440}.
\newblock


\bibitem[\protect\citeauthoryear{Chakraborty, Majumder, Yu, and
  Menzies}{Chakraborty et~al\mbox{.}}{2020}]%
        {chakraborty2020fairway}
\bibfield{author}{\bibinfo{person}{Joymallya Chakraborty},
  \bibinfo{person}{Suvodeep Majumder}, \bibinfo{person}{Zhe Yu}, {and}
  \bibinfo{person}{Tim Menzies}.} \bibinfo{year}{2020}\natexlab{}.
\newblock \showarticletitle{Fairway: A way to build fair ml software}. In
  \bibinfo{booktitle}{\emph{Proceedings of the 28th ACM Joint Meeting on
  European Software Engineering Conference and Symposium on the Foundations of
  Software Engineering}}. \bibinfo{pages}{654--665}.
\newblock


\bibitem[\protect\citeauthoryear{Chen, Zhang, Sarro, and Harman}{Chen
  et~al\mbox{.}}{2022}]%
        {chen2022maat}
\bibfield{author}{\bibinfo{person}{Zhenpeng Chen}, \bibinfo{person}{Jie Zhang},
  \bibinfo{person}{Federica Sarro}, {and} \bibinfo{person}{Mark Harman}.}
  \bibinfo{year}{2022}\natexlab{}.
\newblock \showarticletitle{MAAT: A Novel Ensemble Approach to Addressing
  Fairness and Performance Bugs for Machine Learning Software}. In
  \bibinfo{booktitle}{\emph{The ACM Joint European Software Engineering
  Conference and Symposium on the Foundations of Software Engineering
  (ESEC/FSE)}}.
\newblock


\bibitem[\protect\citeauthoryear{Dastin}{Dastin}{[n.d.]}]%
        {reutersAmazonScraps}
\bibfield{author}{\bibinfo{person}{Jeffrey Dastin}.}
  \bibinfo{year}{[n.d.]}\natexlab{}.
\newblock \bibinfo{title}{{A}mazon scraps secret {A}{I} recruiting tool that
  showed bias against women --- reuters.com}.
\newblock
  \bibinfo{howpublished}{\url{https://www.reuters.com/article/us-amazon-com-jobs-automation-insight/amazon-scraps-secret-ai-recruiting-tool-that-showed-bias-against-women-idUSKCN1MK08G}}.
\newblock
\newblock
\shownote{[Accessed 01-Feb-2023].}


\bibitem[\protect\citeauthoryear{Eberhart and Kennedy}{Eberhart and
  Kennedy}{1995}]%
        {eberhart1995particle}
\bibfield{author}{\bibinfo{person}{Russell Eberhart} {and}
  \bibinfo{person}{James Kennedy}.} \bibinfo{year}{1995}\natexlab{}.
\newblock \showarticletitle{Particle swarm optimization}. In
  \bibinfo{booktitle}{\emph{Proceedings of the IEEE international conference on
  neural networks}}, Vol.~\bibinfo{volume}{4}. Citeseer,
  \bibinfo{pages}{1942--1948}.
\newblock


\bibitem[\protect\citeauthoryear{Fern{\'a}ndez, Garcia, Herrera, and
  Chawla}{Fern{\'a}ndez et~al\mbox{.}}{2018}]%
        {fernandez2018smote}
\bibfield{author}{\bibinfo{person}{Alberto Fern{\'a}ndez},
  \bibinfo{person}{Salvador Garcia}, \bibinfo{person}{Francisco Herrera}, {and}
  \bibinfo{person}{Nitesh~V Chawla}.} \bibinfo{year}{2018}\natexlab{}.
\newblock \showarticletitle{SMOTE for learning from imbalanced data: progress
  and challenges, marking the 15-year anniversary}.
\newblock \bibinfo{journal}{\emph{Journal of artificial intelligence research}}
   \bibinfo{volume}{61} (\bibinfo{year}{2018}), \bibinfo{pages}{863--905}.
\newblock


\bibitem[\protect\citeauthoryear{Fu, Xian, Gao, Zhao, Huang, Ge, Xu, Geng,
  Shah, Zhang, et~al\mbox{.}}{Fu et~al\mbox{.}}{2020}]%
        {fu2020fairness}
\bibfield{author}{\bibinfo{person}{Zuohui Fu}, \bibinfo{person}{Yikun Xian},
  \bibinfo{person}{Ruoyuan Gao}, \bibinfo{person}{Jieyu Zhao},
  \bibinfo{person}{Qiaoying Huang}, \bibinfo{person}{Yingqiang Ge},
  \bibinfo{person}{Shuyuan Xu}, \bibinfo{person}{Shijie Geng},
  \bibinfo{person}{Chirag Shah}, \bibinfo{person}{Yongfeng Zhang},
  {et~al\mbox{.}}} \bibinfo{year}{2020}\natexlab{}.
\newblock \showarticletitle{Fairness-aware explainable recommendation over
  knowledge graphs}. In \bibinfo{booktitle}{\emph{Proceedings of the 43rd
  International ACM SIGIR Conference on Research and Development in Information
  Retrieval}}. \bibinfo{pages}{69--78}.
\newblock


\bibitem[\protect\citeauthoryear{Galhotra, Brun, and Meliou}{Galhotra
  et~al\mbox{.}}{2017}]%
        {galhotra2017fairness}
\bibfield{author}{\bibinfo{person}{Sainyam Galhotra}, \bibinfo{person}{Yuriy
  Brun}, {and} \bibinfo{person}{Alexandra Meliou}.}
  \bibinfo{year}{2017}\natexlab{}.
\newblock \showarticletitle{Fairness testing: testing software for
  discrimination}. In \bibinfo{booktitle}{\emph{Proceedings of the 2017 11th
  Joint meeting on foundations of software engineering}}.
  \bibinfo{pages}{498--510}.
\newblock


\bibitem[\protect\citeauthoryear{Gopnik, Schulz, and Schulz}{Gopnik
  et~al\mbox{.}}{2007}]%
        {gopnik2007causal}
\bibfield{author}{\bibinfo{person}{Alison Gopnik}, \bibinfo{person}{Laura
  Schulz}, {and} \bibinfo{person}{Laura~Elizabeth Schulz}.}
  \bibinfo{year}{2007}\natexlab{}.
\newblock \bibinfo{booktitle}{\emph{Causal learning: Psychology, philosophy,
  and computation}}.
\newblock \bibinfo{publisher}{Oxford University Press}.
\newblock


\bibitem[\protect\citeauthoryear{guofei9987}{guofei9987}{2021}]%
        {misc_Scikit-opt}
\bibfield{author}{\bibinfo{person}{guofei9987}.}
  \bibinfo{year}{2021}\natexlab{}.
\newblock \bibinfo{title}{{Scikit-opt}}.
\newblock \bibinfo{howpublished}{github}.
\newblock
\newblock
\shownote{https://scikit-opt.github.io/.}


\bibitem[\protect\citeauthoryear{Hardt, Price, and Srebro}{Hardt
  et~al\mbox{.}}{2016}]%
        {hardt2016equality}
\bibfield{author}{\bibinfo{person}{Moritz Hardt}, \bibinfo{person}{Eric Price},
  {and} \bibinfo{person}{Nati Srebro}.} \bibinfo{year}{2016}\natexlab{}.
\newblock \showarticletitle{Equality of opportunity in supervised learning}.
\newblock \bibinfo{journal}{\emph{Advances in neural information processing
  systems}}  \bibinfo{volume}{29} (\bibinfo{year}{2016}).
\newblock


\bibitem[\protect\citeauthoryear{Heckman}{Heckman}{2000}]%
        {heckman2000causal}
\bibfield{author}{\bibinfo{person}{James~J Heckman}.}
  \bibinfo{year}{2000}\natexlab{}.
\newblock \showarticletitle{Causal parameters and policy analysis in economics:
  A twentieth century retrospective}.
\newblock \bibinfo{journal}{\emph{The Quarterly Journal of Economics}}
  \bibinfo{volume}{115}, \bibinfo{number}{1} (\bibinfo{year}{2000}),
  \bibinfo{pages}{45--97}.
\newblock


\bibitem[\protect\citeauthoryear{Hettiachchi, Sanderson, Goncalves, Hosio,
  Kazai, Lease, Schaekermann, and Yilmaz}{Hettiachchi et~al\mbox{.}}{2021}]%
        {hettiachchi2021investigating}
\bibfield{author}{\bibinfo{person}{Danula Hettiachchi}, \bibinfo{person}{Mark
  Sanderson}, \bibinfo{person}{Jorge Goncalves}, \bibinfo{person}{Simo Hosio},
  \bibinfo{person}{Gabriella Kazai}, \bibinfo{person}{Matthew Lease},
  \bibinfo{person}{Mike Schaekermann}, {and} \bibinfo{person}{Emine Yilmaz}.}
  \bibinfo{year}{2021}\natexlab{}.
\newblock \showarticletitle{Investigating and Mitigating Biases in Crowdsourced
  Data}. In \bibinfo{booktitle}{\emph{Companion Publication of the 2021
  Conference on Computer Supported Cooperative Work and Social Computing}}.
  \bibinfo{pages}{331--334}.
\newblock


\bibitem[\protect\citeauthoryear{Hort, Zhang, Sarro, and Harman}{Hort
  et~al\mbox{.}}{2021}]%
        {hort2021fairea}
\bibfield{author}{\bibinfo{person}{Max Hort}, \bibinfo{person}{Jie~M Zhang},
  \bibinfo{person}{Federica Sarro}, {and} \bibinfo{person}{Mark Harman}.}
  \bibinfo{year}{2021}\natexlab{}.
\newblock \showarticletitle{Fairea: A model behaviour mutation approach to
  benchmarking bias mitigation methods}. In
  \bibinfo{booktitle}{\emph{Proceedings of the 29th ACM Joint Meeting on
  European Software Engineering Conference and Symposium on the Foundations of
  Software Engineering}}. \bibinfo{pages}{994--1006}.
\newblock


\bibitem[\protect\citeauthoryear{Hube, Fetahu, and Gadiraju}{Hube
  et~al\mbox{.}}{2019}]%
        {hube2019understanding}
\bibfield{author}{\bibinfo{person}{Christoph Hube}, \bibinfo{person}{Besnik
  Fetahu}, {and} \bibinfo{person}{Ujwal Gadiraju}.}
  \bibinfo{year}{2019}\natexlab{}.
\newblock \showarticletitle{Understanding and mitigating worker biases in the
  crowdsourced collection of subjective judgments}. In
  \bibinfo{booktitle}{\emph{Proceedings of the 2019 CHI Conference on Human
  Factors in Computing Systems}}. \bibinfo{pages}{1--12}.
\newblock


\bibitem[\protect\citeauthoryear{Imbens and Rubin}{Imbens and Rubin}{2015}]%
        {imbens2015causal}
\bibfield{author}{\bibinfo{person}{Guido~W Imbens} {and}
  \bibinfo{person}{Donald~B Rubin}.} \bibinfo{year}{2015}\natexlab{}.
\newblock \bibinfo{booktitle}{\emph{Causal inference in statistics, social, and
  biomedical sciences}}.
\newblock \bibinfo{publisher}{Cambridge University Press}.
\newblock


\bibitem[\protect\citeauthoryear{Janai, G{\"u}ney, Behl, Geiger,
  et~al\mbox{.}}{Janai et~al\mbox{.}}{2020}]%
        {janai2020computer}
\bibfield{author}{\bibinfo{person}{Joel Janai}, \bibinfo{person}{Fatma
  G{\"u}ney}, \bibinfo{person}{Aseem Behl}, \bibinfo{person}{Andreas Geiger},
  {et~al\mbox{.}}} \bibinfo{year}{2020}\natexlab{}.
\newblock \showarticletitle{Computer vision for autonomous vehicles: Problems,
  datasets and state of the art}.
\newblock \bibinfo{journal}{\emph{Foundations and Trends{\textregistered} in
  Computer Graphics and Vision}} \bibinfo{volume}{12}, \bibinfo{number}{1--3}
  (\bibinfo{year}{2020}), \bibinfo{pages}{1--308}.
\newblock


\bibitem[\protect\citeauthoryear{Jobin, Ienca, and Vayena}{Jobin
  et~al\mbox{.}}{2019}]%
        {jobin2019global}
\bibfield{author}{\bibinfo{person}{Anna Jobin}, \bibinfo{person}{Marcello
  Ienca}, {and} \bibinfo{person}{Effy Vayena}.}
  \bibinfo{year}{2019}\natexlab{}.
\newblock \showarticletitle{The global landscape of AI ethics guidelines}.
\newblock \bibinfo{journal}{\emph{Nature Machine Intelligence}}
  \bibinfo{volume}{1}, \bibinfo{number}{9} (\bibinfo{year}{2019}),
  \bibinfo{pages}{389--399}.
\newblock


\bibitem[\protect\citeauthoryear{Kamiran and Calders}{Kamiran and
  Calders}{2012}]%
        {kamiran2012data}
\bibfield{author}{\bibinfo{person}{Faisal Kamiran} {and} \bibinfo{person}{Toon
  Calders}.} \bibinfo{year}{2012}\natexlab{}.
\newblock \showarticletitle{Data preprocessing techniques for classification
  without discrimination}.
\newblock \bibinfo{journal}{\emph{Knowledge and information systems}}
  \bibinfo{volume}{33}, \bibinfo{number}{1} (\bibinfo{year}{2012}),
  \bibinfo{pages}{1--33}.
\newblock


\bibitem[\protect\citeauthoryear{Kamiran, Karim, and Zhang}{Kamiran
  et~al\mbox{.}}{2012}]%
        {kamiran2012decision}
\bibfield{author}{\bibinfo{person}{Faisal Kamiran}, \bibinfo{person}{Asim
  Karim}, {and} \bibinfo{person}{Xiangliang Zhang}.}
  \bibinfo{year}{2012}\natexlab{}.
\newblock \showarticletitle{Decision theory for discrimination-aware
  classification}. In \bibinfo{booktitle}{\emph{2012 IEEE 12th International
  Conference on Data Mining}}. IEEE, \bibinfo{pages}{924--929}.
\newblock


\bibitem[\protect\citeauthoryear{Kamishima, Akaho, Asoh, and Sakuma}{Kamishima
  et~al\mbox{.}}{2012}]%
        {kamishima2012fairness}
\bibfield{author}{\bibinfo{person}{Toshihiro Kamishima},
  \bibinfo{person}{Shotaro Akaho}, \bibinfo{person}{Hideki Asoh}, {and}
  \bibinfo{person}{Jun Sakuma}.} \bibinfo{year}{2012}\natexlab{}.
\newblock \showarticletitle{Fairness-aware classifier with prejudice remover
  regularizer}. In \bibinfo{booktitle}{\emph{Joint European conference on
  machine learning and knowledge discovery in databases}}. Springer,
  \bibinfo{pages}{35--50}.
\newblock


\bibitem[\protect\citeauthoryear{Kohn and Shavell}{Kohn and Shavell}{1974}]%
        {kohn1974theory}
\bibfield{author}{\bibinfo{person}{Meir~G Kohn} {and} \bibinfo{person}{Steven
  Shavell}.} \bibinfo{year}{1974}\natexlab{}.
\newblock \showarticletitle{The theory of search}.
\newblock \bibinfo{journal}{\emph{Journal of Economic Theory}}
  \bibinfo{volume}{9}, \bibinfo{number}{2} (\bibinfo{year}{1974}),
  \bibinfo{pages}{93--123}.
\newblock


\bibitem[\protect\citeauthoryear{Kotary, Fioretto, Van~Hentenryck, and
  Zhu}{Kotary et~al\mbox{.}}{2022}]%
        {kotary2022end}
\bibfield{author}{\bibinfo{person}{James Kotary}, \bibinfo{person}{Ferdinando
  Fioretto}, \bibinfo{person}{Pascal Van~Hentenryck}, {and}
  \bibinfo{person}{Ziwei Zhu}.} \bibinfo{year}{2022}\natexlab{}.
\newblock \showarticletitle{End-to-End Learning for Fair Ranking Systems}. In
  \bibinfo{booktitle}{\emph{Proceedings of the ACM Web Conference 2022}}.
  \bibinfo{pages}{3520--3530}.
\newblock


\bibitem[\protect\citeauthoryear{Kuipers and Kassirer}{Kuipers and
  Kassirer}{1984}]%
        {kuipers1984causal}
\bibfield{author}{\bibinfo{person}{Benjamin Kuipers} {and}
  \bibinfo{person}{Jerome~P Kassirer}.} \bibinfo{year}{1984}\natexlab{}.
\newblock \showarticletitle{Causal reasoning in medicine: analysis of a
  protocol}.
\newblock \bibinfo{journal}{\emph{Cognitive Science}} \bibinfo{volume}{8},
  \bibinfo{number}{4} (\bibinfo{year}{1984}), \bibinfo{pages}{363--385}.
\newblock


\bibitem[\protect\citeauthoryear{Lahoti, Beutel, Chen, Lee, Prost, Thain, Wang,
  and Chi}{Lahoti et~al\mbox{.}}{2020}]%
        {lahoti2020fairness}
\bibfield{author}{\bibinfo{person}{Preethi Lahoti}, \bibinfo{person}{Alex
  Beutel}, \bibinfo{person}{Jilin Chen}, \bibinfo{person}{Kang Lee},
  \bibinfo{person}{Flavien Prost}, \bibinfo{person}{Nithum Thain},
  \bibinfo{person}{Xuezhi Wang}, {and} \bibinfo{person}{Ed Chi}.}
  \bibinfo{year}{2020}\natexlab{}.
\newblock \showarticletitle{Fairness without demographics through adversarially
  reweighted learning}.
\newblock \bibinfo{journal}{\emph{Advances in neural information processing
  systems}}  \bibinfo{volume}{33} (\bibinfo{year}{2020}),
  \bibinfo{pages}{728--740}.
\newblock


\bibitem[\protect\citeauthoryear{LeCun, Bengio, and Hinton}{LeCun
  et~al\mbox{.}}{2015}]%
        {lecun2015deep}
\bibfield{author}{\bibinfo{person}{Yann LeCun}, \bibinfo{person}{Yoshua
  Bengio}, {and} \bibinfo{person}{Geoffrey Hinton}.}
  \bibinfo{year}{2015}\natexlab{}.
\newblock \showarticletitle{Deep learning}.
\newblock \bibinfo{journal}{\emph{nature}} \bibinfo{volume}{521},
  \bibinfo{number}{7553} (\bibinfo{year}{2015}), \bibinfo{pages}{436--444}.
\newblock


\bibitem[\protect\citeauthoryear{Li, Chen, Fu, Ge, and Zhang}{Li
  et~al\mbox{.}}{2021}]%
        {li2021user}
\bibfield{author}{\bibinfo{person}{Yunqi Li}, \bibinfo{person}{Hanxiong Chen},
  \bibinfo{person}{Zuohui Fu}, \bibinfo{person}{Yingqiang Ge}, {and}
  \bibinfo{person}{Yongfeng Zhang}.} \bibinfo{year}{2021}\natexlab{}.
\newblock \showarticletitle{User-oriented fairness in recommendation}. In
  \bibinfo{booktitle}{\emph{Proceedings of the Web Conference 2021}}.
  \bibinfo{pages}{624--632}.
\newblock


\bibitem[\protect\citeauthoryear{Madaio, Stark, Wortman~Vaughan, and
  Wallach}{Madaio et~al\mbox{.}}{2020}]%
        {madaio2020co}
\bibfield{author}{\bibinfo{person}{Michael~A Madaio}, \bibinfo{person}{Luke
  Stark}, \bibinfo{person}{Jennifer Wortman~Vaughan}, {and}
  \bibinfo{person}{Hanna Wallach}.} \bibinfo{year}{2020}\natexlab{}.
\newblock \showarticletitle{Co-designing checklists to understand
  organizational challenges and opportunities around fairness in AI}. In
  \bibinfo{booktitle}{\emph{Proceedings of the 2020 CHI Conference on Human
  Factors in Computing Systems}}. \bibinfo{pages}{1--14}.
\newblock


\bibitem[\protect\citeauthoryear{Mann and Whitney}{Mann and Whitney}{1947}]%
        {mann1947test}
\bibfield{author}{\bibinfo{person}{Henry~B Mann} {and}
  \bibinfo{person}{Donald~R Whitney}.} \bibinfo{year}{1947}\natexlab{}.
\newblock \showarticletitle{On a test of whether one of two random variables is
  stochastically larger than the other}.
\newblock \bibinfo{journal}{\emph{The annals of mathematical statistics}}
  (\bibinfo{year}{1947}), \bibinfo{pages}{50--60}.
\newblock


\bibitem[\protect\citeauthoryear{Mehrabi, Morstatter, Saxena, Lerman, and
  Galstyan}{Mehrabi et~al\mbox{.}}{2021}]%
        {mehrabi2021survey}
\bibfield{author}{\bibinfo{person}{Ninareh Mehrabi}, \bibinfo{person}{Fred
  Morstatter}, \bibinfo{person}{Nripsuta Saxena}, \bibinfo{person}{Kristina
  Lerman}, {and} \bibinfo{person}{Aram Galstyan}.}
  \bibinfo{year}{2021}\natexlab{}.
\newblock \showarticletitle{A survey on bias and fairness in machine learning}.
\newblock \bibinfo{journal}{\emph{ACM Computing Surveys (CSUR)}}
  \bibinfo{volume}{54}, \bibinfo{number}{6} (\bibinfo{year}{2021}),
  \bibinfo{pages}{1--35}.
\newblock


\bibitem[\protect\citeauthoryear{Mohammed, Rawashdeh, and Abdullah}{Mohammed
  et~al\mbox{.}}{2020}]%
        {mohammed2020machine}
\bibfield{author}{\bibinfo{person}{Roweida Mohammed}, \bibinfo{person}{Jumanah
  Rawashdeh}, {and} \bibinfo{person}{Malak Abdullah}.}
  \bibinfo{year}{2020}\natexlab{}.
\newblock \showarticletitle{Machine learning with oversampling and
  undersampling techniques: overview study and experimental results}. In
  \bibinfo{booktitle}{\emph{2020 11th international conference on information
  and communication systems (ICICS)}}. IEEE, \bibinfo{pages}{243--248}.
\newblock


\bibitem[\protect\citeauthoryear{Moro}{Moro}{2012}]%
        {misc_bank_marketing_222}
\bibfield{author}{\bibinfo{person}{Rita P. \& Cortez~P. Moro, S.}}
  \bibinfo{year}{2012}\natexlab{}.
\newblock \bibinfo{title}{{The Bank Marketing dateset}}.
\newblock \bibinfo{howpublished}{UCI Machine Learning Repository}.
\newblock
\newblock
\shownote{https://archive.ics.uci.edu/ml/datasets/bank+marketing.}


\bibitem[\protect\citeauthoryear{Noble}{Noble}{2006}]%
        {noble2006support}
\bibfield{author}{\bibinfo{person}{William~S Noble}.}
  \bibinfo{year}{2006}\natexlab{}.
\newblock \showarticletitle{What is a support vector machine?}
\newblock \bibinfo{journal}{\emph{Nature biotechnology}} \bibinfo{volume}{24},
  \bibinfo{number}{12} (\bibinfo{year}{2006}), \bibinfo{pages}{1565--1567}.
\newblock


\bibitem[\protect\citeauthoryear{Pearl}{Pearl}{2009}]%
        {pearl2009causality}
\bibfield{author}{\bibinfo{person}{Judea Pearl}.}
  \bibinfo{year}{2009}\natexlab{}.
\newblock \bibinfo{booktitle}{\emph{Causality}}.
\newblock \bibinfo{publisher}{Cambridge university press}.
\newblock


\bibitem[\protect\citeauthoryear{Pearl and Mackenzie}{Pearl and
  Mackenzie}{2018}]%
        {pearl2018book}
\bibfield{author}{\bibinfo{person}{Judea Pearl} {and} \bibinfo{person}{Dana
  Mackenzie}.} \bibinfo{year}{2018}\natexlab{}.
\newblock \bibinfo{booktitle}{\emph{The book of why: the new science of cause
  and effect}}.
\newblock \bibinfo{publisher}{Basic books}.
\newblock


\bibitem[\protect\citeauthoryear{Pedregosa, Varoquaux, Gramfort, Michel,
  Thirion, Grisel, Blondel, Prettenhofer, Weiss, Dubourg, Vanderplas, Passos,
  Cournapeau, Brucher, Perrot, and Duchesnay}{Pedregosa et~al\mbox{.}}{2011}]%
        {scikit-learn}
\bibfield{author}{\bibinfo{person}{F. Pedregosa}, \bibinfo{person}{G.
  Varoquaux}, \bibinfo{person}{A. Gramfort}, \bibinfo{person}{V. Michel},
  \bibinfo{person}{B. Thirion}, \bibinfo{person}{O. Grisel},
  \bibinfo{person}{M. Blondel}, \bibinfo{person}{P. Prettenhofer},
  \bibinfo{person}{R. Weiss}, \bibinfo{person}{V. Dubourg}, \bibinfo{person}{J.
  Vanderplas}, \bibinfo{person}{A. Passos}, \bibinfo{person}{D. Cournapeau},
  \bibinfo{person}{M. Brucher}, \bibinfo{person}{M. Perrot}, {and}
  \bibinfo{person}{E. Duchesnay}.} \bibinfo{year}{2011}\natexlab{}.
\newblock \showarticletitle{Scikit-learn: Machine Learning in {P}ython}.
\newblock \bibinfo{journal}{\emph{Journal of Machine Learning Research}}
  \bibinfo{volume}{12} (\bibinfo{year}{2011}), \bibinfo{pages}{2825--2830}.
\newblock


\bibitem[\protect\citeauthoryear{Peng, Chakraborty, and Menzies}{Peng
  et~al\mbox{.}}{2021}]%
        {peng2021xfair}
\bibfield{author}{\bibinfo{person}{Kewen Peng}, \bibinfo{person}{Joymallya
  Chakraborty}, {and} \bibinfo{person}{Tim Menzies}.}
  \bibinfo{year}{2021}\natexlab{}.
\newblock \showarticletitle{xFAIR: Better Fairness via Model-based Rebalancing
  of Protected Attributes}.
\newblock \bibinfo{journal}{\emph{arXiv preprint arXiv:2110.01109}}
  (\bibinfo{year}{2021}).
\newblock


\bibitem[\protect\citeauthoryear{Pleiss, Raghavan, Wu, Kleinberg, and
  Weinberger}{Pleiss et~al\mbox{.}}{2017}]%
        {pleiss2017fairness}
\bibfield{author}{\bibinfo{person}{Geoff Pleiss}, \bibinfo{person}{Manish
  Raghavan}, \bibinfo{person}{Felix Wu}, \bibinfo{person}{Jon Kleinberg}, {and}
  \bibinfo{person}{Kilian~Q Weinberger}.} \bibinfo{year}{2017}\natexlab{}.
\newblock \showarticletitle{On fairness and calibration}.
\newblock \bibinfo{journal}{\emph{Advances in neural information processing
  systems}}  \bibinfo{volume}{30} (\bibinfo{year}{2017}).
\newblock


\bibitem[\protect\citeauthoryear{Prost, Qian, Chen, Chi, Chen, and
  Beutel}{Prost et~al\mbox{.}}{2019}]%
        {prost2019toward}
\bibfield{author}{\bibinfo{person}{Flavien Prost}, \bibinfo{person}{Hai Qian},
  \bibinfo{person}{Qiuwen Chen}, \bibinfo{person}{Ed~H Chi},
  \bibinfo{person}{Jilin Chen}, {and} \bibinfo{person}{Alex Beutel}.}
  \bibinfo{year}{2019}\natexlab{}.
\newblock \showarticletitle{Toward a better trade-off between performance and
  fairness with kernel-based distribution matching}.
\newblock \bibinfo{journal}{\emph{arXiv preprint arXiv:1910.11779}}
  (\bibinfo{year}{2019}).
\newblock


\bibitem[\protect\citeauthoryear{Rastegarpanah, Gummadi, and
  Crovella}{Rastegarpanah et~al\mbox{.}}{2019}]%
        {rastegarpanah2019fighting}
\bibfield{author}{\bibinfo{person}{Bashir Rastegarpanah},
  \bibinfo{person}{Krishna~P Gummadi}, {and} \bibinfo{person}{Mark Crovella}.}
  \bibinfo{year}{2019}\natexlab{}.
\newblock \showarticletitle{Fighting fire with fire: Using antidote data to
  improve polarization and fairness of recommender systems}. In
  \bibinfo{booktitle}{\emph{Proceedings of the twelfth ACM international
  conference on web search and data mining}}. \bibinfo{pages}{231--239}.
\newblock


\bibitem[\protect\citeauthoryear{Sambasivan, Kapania, Highfill, Akrong,
  Paritosh, and Aroyo}{Sambasivan et~al\mbox{.}}{2021}]%
        {sambasivan2021everyone}
\bibfield{author}{\bibinfo{person}{Nithya Sambasivan}, \bibinfo{person}{Shivani
  Kapania}, \bibinfo{person}{Hannah Highfill}, \bibinfo{person}{Diana Akrong},
  \bibinfo{person}{Praveen Paritosh}, {and} \bibinfo{person}{Lora~M Aroyo}.}
  \bibinfo{year}{2021}\natexlab{}.
\newblock \showarticletitle{“Everyone wants to do the model work, not the
  data work”: Data Cascades in High-Stakes AI}. In
  \bibinfo{booktitle}{\emph{proceedings of the 2021 CHI Conference on Human
  Factors in Computing Systems}}. \bibinfo{pages}{1--15}.
\newblock


\bibitem[\protect\citeauthoryear{Schwarting, Alonso-Mora, and Rus}{Schwarting
  et~al\mbox{.}}{2018}]%
        {schwarting2018planning}
\bibfield{author}{\bibinfo{person}{Wilko Schwarting}, \bibinfo{person}{Javier
  Alonso-Mora}, {and} \bibinfo{person}{Daniela Rus}.}
  \bibinfo{year}{2018}\natexlab{}.
\newblock \showarticletitle{Planning and decision-making for autonomous
  vehicles}.
\newblock \bibinfo{journal}{\emph{Annual Review of Control, Robotics, and
  Autonomous Systems}}  \bibinfo{volume}{1} (\bibinfo{year}{2018}),
  \bibinfo{pages}{187--210}.
\newblock


\bibitem[\protect\citeauthoryear{Sheng and Zhang}{Sheng and Zhang}{2019}]%
        {sheng2019machine}
\bibfield{author}{\bibinfo{person}{Victor~S Sheng} {and} \bibinfo{person}{Jing
  Zhang}.} \bibinfo{year}{2019}\natexlab{}.
\newblock \showarticletitle{Machine learning with crowdsourcing: A brief
  summary of the past research and future directions}. In
  \bibinfo{booktitle}{\emph{Proceedings of the AAAI conference on artificial
  intelligence}}, Vol.~\bibinfo{volume}{33}. \bibinfo{pages}{9837--9843}.
\newblock


\bibitem[\protect\citeauthoryear{Sun, Sun, Pham, and Shi}{Sun
  et~al\mbox{.}}{2022}]%
        {sun2022causality}
\bibfield{author}{\bibinfo{person}{Bing Sun}, \bibinfo{person}{Jun Sun},
  \bibinfo{person}{Long~H Pham}, {and} \bibinfo{person}{Jie Shi}.}
  \bibinfo{year}{2022}\natexlab{}.
\newblock \showarticletitle{Causality-based neural network repair}. In
  \bibinfo{booktitle}{\emph{Proceedings of the 44th International Conference on
  Software Engineering}}. \bibinfo{pages}{338--349}.
\newblock


\bibitem[\protect\citeauthoryear{Tsintzou, Pitoura, and Tsaparas}{Tsintzou
  et~al\mbox{.}}{2018}]%
        {tsintzou2018bias}
\bibfield{author}{\bibinfo{person}{Virginia Tsintzou},
  \bibinfo{person}{Evaggelia Pitoura}, {and} \bibinfo{person}{Panayiotis
  Tsaparas}.} \bibinfo{year}{2018}\natexlab{}.
\newblock \showarticletitle{Bias disparity in recommendation systems}.
\newblock \bibinfo{journal}{\emph{arXiv preprint arXiv:1811.01461}}
  (\bibinfo{year}{2018}).
\newblock


\bibitem[\protect\citeauthoryear{Tsioutsiouliklis, Pitoura, Semertzidis, and
  Tsaparas}{Tsioutsiouliklis et~al\mbox{.}}{2022}]%
        {tsioutsiouliklis2022link}
\bibfield{author}{\bibinfo{person}{Sotiris Tsioutsiouliklis},
  \bibinfo{person}{Evaggelia Pitoura}, \bibinfo{person}{Konstantinos
  Semertzidis}, {and} \bibinfo{person}{Panayiotis Tsaparas}.}
  \bibinfo{year}{2022}\natexlab{}.
\newblock \showarticletitle{Link Recommendations for PageRank Fairness}. In
  \bibinfo{booktitle}{\emph{Proceedings of the ACM Web Conference 2022}}.
  \bibinfo{pages}{3541--3551}.
\newblock


\bibitem[\protect\citeauthoryear{Udeshi, Arora, and Chattopadhyay}{Udeshi
  et~al\mbox{.}}{2018}]%
        {udeshi2018automated}
\bibfield{author}{\bibinfo{person}{Sakshi Udeshi}, \bibinfo{person}{Pryanshu
  Arora}, {and} \bibinfo{person}{Sudipta Chattopadhyay}.}
  \bibinfo{year}{2018}\natexlab{}.
\newblock \showarticletitle{Automated directed fairness testing}. In
  \bibinfo{booktitle}{\emph{Proceedings of the 33rd ACM/IEEE International
  Conference on Automated Software Engineering}}. \bibinfo{pages}{98--108}.
\newblock


\bibitem[\protect\citeauthoryear{Van~Laarhoven and Aarts}{Van~Laarhoven and
  Aarts}{1987}]%
        {van1987simulated}
\bibfield{author}{\bibinfo{person}{Peter~JM Van~Laarhoven} {and}
  \bibinfo{person}{Emile~HL Aarts}.} \bibinfo{year}{1987}\natexlab{}.
\newblock \showarticletitle{Simulated annealing}.
\newblock In \bibinfo{booktitle}{\emph{Simulated annealing: Theory and
  applications}}. \bibinfo{publisher}{Springer}, \bibinfo{pages}{7--15}.
\newblock


\bibitem[\protect\citeauthoryear{Varian}{Varian}{2016}]%
        {varian2016causal}
\bibfield{author}{\bibinfo{person}{Hal~R Varian}.}
  \bibinfo{year}{2016}\natexlab{}.
\newblock \showarticletitle{Causal inference in economics and marketing}.
\newblock \bibinfo{journal}{\emph{Proceedings of the National Academy of
  Sciences}} \bibinfo{volume}{113}, \bibinfo{number}{27}
  (\bibinfo{year}{2016}), \bibinfo{pages}{7310--7315}.
\newblock


\bibitem[\protect\citeauthoryear{Voigt and Von~dem Bussche}{Voigt and Von~dem
  Bussche}{2017}]%
        {voigt2017eu}
\bibfield{author}{\bibinfo{person}{Paul Voigt} {and} \bibinfo{person}{Axel
  Von~dem Bussche}.} \bibinfo{year}{2017}\natexlab{}.
\newblock \showarticletitle{The eu general data protection regulation (gdpr)}.
\newblock \bibinfo{journal}{\emph{A Practical Guide, 1st Ed., Cham: Springer
  International Publishing}} \bibinfo{volume}{10}, \bibinfo{number}{3152676}
  (\bibinfo{year}{2017}), \bibinfo{pages}{10--5555}.
\newblock


\bibitem[\protect\citeauthoryear{Whitley}{Whitley}{1994}]%
        {whitley1994genetic}
\bibfield{author}{\bibinfo{person}{Darrell Whitley}.}
  \bibinfo{year}{1994}\natexlab{}.
\newblock \showarticletitle{A genetic algorithm tutorial}.
\newblock \bibinfo{journal}{\emph{Statistics and computing}}
  \bibinfo{volume}{4}, \bibinfo{number}{2} (\bibinfo{year}{1994}),
  \bibinfo{pages}{65--85}.
\newblock


\bibitem[\protect\citeauthoryear{Wright}{Wright}{1995}]%
        {wright1995logistic}
\bibfield{author}{\bibinfo{person}{Raymond~E Wright}.}
  \bibinfo{year}{1995}\natexlab{}.
\newblock \showarticletitle{Logistic regression.}
\newblock  (\bibinfo{year}{1995}).
\newblock


\bibitem[\protect\citeauthoryear{Wu, Wu, Wang, Huang, and Xie}{Wu
  et~al\mbox{.}}{2021}]%
        {wu2021fairness}
\bibfield{author}{\bibinfo{person}{Chuhan Wu}, \bibinfo{person}{Fangzhao Wu},
  \bibinfo{person}{Xiting Wang}, \bibinfo{person}{Yongfeng Huang}, {and}
  \bibinfo{person}{Xing Xie}.} \bibinfo{year}{2021}\natexlab{}.
\newblock \showarticletitle{Fairness-aware news recommendation with decomposed
  adversarial learning}. In \bibinfo{booktitle}{\emph{Proceedings of the AAAI
  Conference on Artificial Intelligence}}, Vol.~\bibinfo{volume}{35}.
  \bibinfo{pages}{4462--4469}.
\newblock


\bibitem[\protect\citeauthoryear{Zemel, Wu, Swersky, Pitassi, and Dwork}{Zemel
  et~al\mbox{.}}{2013}]%
        {zemel2013learning}
\bibfield{author}{\bibinfo{person}{Rich Zemel}, \bibinfo{person}{Yu Wu},
  \bibinfo{person}{Kevin Swersky}, \bibinfo{person}{Toni Pitassi}, {and}
  \bibinfo{person}{Cynthia Dwork}.} \bibinfo{year}{2013}\natexlab{}.
\newblock \showarticletitle{Learning fair representations}. In
  \bibinfo{booktitle}{\emph{International conference on machine learning}}.
  PMLR, \bibinfo{pages}{325--333}.
\newblock


\bibitem[\protect\citeauthoryear{Zhang, Lemoine, and Mitchell}{Zhang
  et~al\mbox{.}}{2018}]%
        {zhang2018mitigating}
\bibfield{author}{\bibinfo{person}{Brian~Hu Zhang}, \bibinfo{person}{Blake
  Lemoine}, {and} \bibinfo{person}{Margaret Mitchell}.}
  \bibinfo{year}{2018}\natexlab{}.
\newblock \showarticletitle{Mitigating unwanted biases with adversarial
  learning}. In \bibinfo{booktitle}{\emph{Proceedings of the 2018 AAAI/ACM
  Conference on AI, Ethics, and Society}}. \bibinfo{pages}{335--340}.
\newblock


\bibitem[\protect\citeauthoryear{Zhang and Harman}{Zhang and Harman}{2021}]%
        {zhang2021ignorance}
\bibfield{author}{\bibinfo{person}{Jie~M Zhang} {and} \bibinfo{person}{Mark
  Harman}.} \bibinfo{year}{2021}\natexlab{}.
\newblock \showarticletitle{" Ignorance and Prejudice" in Software Fairness}.
  In \bibinfo{booktitle}{\emph{2021 IEEE/ACM 43rd International Conference on
  Software Engineering (ICSE)}}. IEEE, \bibinfo{pages}{1436--1447}.
\newblock


\bibitem[\protect\citeauthoryear{Zhang, Harman, Ma, and Liu}{Zhang
  et~al\mbox{.}}{2020}]%
        {zhang2020machine}
\bibfield{author}{\bibinfo{person}{Jie~M Zhang}, \bibinfo{person}{Mark Harman},
  \bibinfo{person}{Lei Ma}, {and} \bibinfo{person}{Yang Liu}.}
  \bibinfo{year}{2020}\natexlab{}.
\newblock \showarticletitle{Machine learning testing: Survey, landscapes and
  horizons}.
\newblock \bibinfo{journal}{\emph{IEEE Transactions on Software Engineering}}
  (\bibinfo{year}{2020}).
\newblock


\end{thebibliography}
